\documentclass[10pt,twocolumn,letterpaper]{article}
\usepackage{cvpr}              
\usepackage{amsmath}
\usepackage{amssymb}
\usepackage{nicefrac}       
\usepackage{thm-restate}
\usepackage{amsmath,amsfonts,bm}
\usepackage{amssymb,amsmath}
\usepackage{amsthm}
\usepackage[scr=boondoxo]{mathalfa}
\usepackage[pagebackref,breaklinks,colorlinks]{hyperref}
\usepackage[capitalize]{cleveref}
\crefname{section}{Sec.}{Secs.}
\Crefname{section}{Section}{Sections}
\Crefname{table}{Table}{Tables}
\crefname{table}{Tab.}{Tabs.}
\usepackage{comment}
\usepackage{color}
\usepackage{graphicx}
\usepackage{booktabs} 
\usepackage{hyperref}

\usepackage{mathtools}
\usepackage{breqn}
\usepackage{pgfplots}
\usepackage[utf8]{inputenc}
\usepackage{microtype}

\usepackage{pifont}
\usepackage[ruled,vlined]{algorithm2e}
\usepackage{float}
\newfloat{eqn}{htbp}{loe}
\floatname{eqn}{Equation}
\usepackage{tabularx}
\usepackage[nopar]{lipsum}
\usepackage{changepage}
\usepackage{multirow}
\usepackage{multicol}
\definecolor{smallcnn}{RGB}{85,163,205}
\definecolor{resnet9}{RGB}{73,84,176}
\definecolor{resnet18}{RGB}{59,33,39}
\definecolor{resnet34}{RGB}{156,47,69}
\definecolor{resnet50}{RGB}{233,111,54}
\definecolor{wide_resnet50_2}{RGB}{27,97,69}
\definecolor{resnet101}{RGB}{105,123,48}
\definecolor{wide_resnet101_2}{RGB}{200,123,124}
\definecolor{resnet152}{RGB}{205,162,224}
\definecolor{vgg}{RGB}{198,225,241}
\definecolor{good_ASR}{RGB}{233,141,107}
\definecolor{bad_ASR}{RGB}{108,43,109}
\definecolor{dark-green}{rgb}{0,0.6,0}
\definecolor{amaranth}{rgb}{0.9, 0.17, 0.31}
\definecolor{bostonuniversityred}{rgb}{0.8, 0.0, 0.0}
\definecolor{brightpink}{rgb}{1.0, 0.0, 0.5}
\definecolor{darklava}{rgb}{0.28, 0.24, 0.2}
\definecolor{darkgreen}{rgb}{0.0, 0.2, 0.13}
\definecolor{coolblack}{rgb}{0.0, 0.18, 0.39}
\definecolor{blue-violet}{rgb}{0.54, 0.17, 0.89}
\definecolor{caribbeangreen}{rgb}{0.0, 0.8, 0.6}
\definecolor{cobalt}{rgb}{0.0, 0.28, 0.67}
\definecolor{darkcyan}{rgb}{0.0, 0.55, 0.55}
\definecolor{darkmidnightblue}{rgb}{0.0, 0.2, 0.4}
\definecolor{darkslateblue}{rgb}{0.28, 0.24, 0.55}
\definecolor{electricultramarine}{rgb}{0.25, 0.0, 1.0}
\usepackage{tikz}
\newcommand*\circled[1]{\tikz[baseline=(char.base)]{
            \node[shape=circle,draw,inner sep=0.5pt] (char) {#1};}}

\usepackage{cases}
\usepackage{enumitem}
\DeclareMathOperator*{\argmax}{arg\,max}

\usepackage{wrapfig}
\usepackage[square,sort,comma,numbers]{natbib}
\usepackage[scaled=.7]{beramono}
\usepackage[T1]{fontenc}  
\usepackage{collectbox}
\makeatletter

\def\cvprPaperID{*****} 
\def\confName{CVPR}
\def\confYear{2022}

\begin{document}

\title{Low-Loss Subspace Compression for \\ Clean Gains against Multi-Agent Backdoor Attacks} 

\author{Siddhartha Datta\\
University of Oxford\\
{\tt\small siddhartha.datta@cs.ox.ac.uk}
\and
Nigel Shadbolt\\
University of Oxford\\
{\tt\small nigel.shadbolt@cs.ox.ac.uk}
}

\maketitle

\begin{abstract}
Recent exploration of the multi-agent backdoor attack demonstrated the \textit{backfiring effect}, a natural defense against backdoor attacks where backdoored inputs are randomly classified. This yields a side-effect of low accuracy w.r.t. clean labels, which motivates this paper's work on the construction of multi-agent backdoor defenses that maximize accuracy w.r.t. clean labels and minimize that of poison labels. Founded upon agent dynamics and low-loss subspace construction, we contribute three defenses that yield improved multi-agent backdoor robustness.
\end{abstract}

\section{Introduction}

In many practical scenarios, collaborative learning and outsourced data collection training regimes are adopted to scale-up model accuracy, certainty and diversity. 
These settings can thus be prone to malicious agents with the intent of compromising such models, one exploit being known as a backdoor attack \cite{gu2019badnets}.
A subsequent string of attacks and defenses have been studied to counter the single-agent backdoor attack \cite{gao2020backdoor, li2021backdoor}, where a single malicious agent inserts train-time perturbations into the joint dataset such that the perturbation at test-time yields a specific target label.

Datta \& Shadbolt (2022) \cite{datta2022backdoors} evaluated the multi-agent backdoor attack, where multiple agents fail to mount a successful attack. Despite a broad enumeration of the different attack configurations, including simulated cooperation, stylization, or escalation, if the defender opts to use a defense, then the equilibrium accuracy with respect to (w.r.t.) poisoned labels tends to be $\frac{1}{|\mathcal{Y}|}$.
They denote this phenomenon as the \textit{backfiring effect}, and explain that in the presence of a backdoor perturbation, the model randomly samples the predicted label.

Further inspecting the multi-agent backdoor attack, 
we find that one of the downsides of the \textit{backfiring effect} (though not in conflict with the objectives of backdoor defense) is that the accuracy w.r.t. clean labels also tends to be low and approximates $\frac{1}{|Y|}$ if the input contains the backdoor perturbation.
In this work, we evaluate existing single-agent backdoor defenses to find that they cannot increase the accuracy w.r.t. clean labels during a backdoor attack, and furthermore contribute three multi-agent backdoor defenses motivated by agent dynamics. 
In addition, we are among the first to construct multi-distributional subspaces to tackle joint distribution shifts.
We hope our contribution can push for robust model deployment in practical settings, and counter backdoor defense considerations previously not considered.

\section{Preliminaries \& Related Work}
\label{2}

In single-agent backdoor attacks~\citep{gu2019badnets}, the attack objective is to maximize the attack success rate in the presence of the trigger while retaining the  accuracy of the model on clean samples.
To achieve this attack objective, a common attack vector is through that of outsourced data collection~\citep{gu2019badnets, chen2017targeted, shafahi2018poison, zhu2019transferable, saha2020hidden, 9230411, dattaSelf}.
Backdoor attacks~\citep{48698, NEURIPS2020_b8ffa41d, pmlr-v108-bagdasaryan20a, huang2020dynamic} 
and poisoning attacks~\citep{NEURIPS2018_331316d4, mahloujifar2018multiparty, pmlr-v97-mahloujifar19a, CHEN2021100002, 247652} 
against federated learning systems and against multi-party learning models
have been demonstrated, but with a single attacker intending to compromise multiple victims (i.e. single attacker vs multiple defenders);
for example, with a single attacker controlling multiple participant nodes in the federated learning setup \citep{pmlr-v108-bagdasaryan20a};
or decomposing a backdoor trigger pattern into multiple distributed small patterns to be injected by multiple participant nodes controlled by a single attacker \citep{Xie2020DBA}.
Datta \& Shadbolt (2022) \cite{datta2022backdoors} evaluated the multi-agent backdoor attack, to evaluate multiple attackers against each other and a defender, and the attack objective is individualized for each attacker. 
Though some upsides presented in the work include a natural robustness to backdoor attackers when more than one attacker attempts to backdoor a model (resulting in collective minimization of attack success rate attributed to the \textit{backfiring effect}), 
further inspection of the setup leads us to believe there are remaining aspects of robustness to be addressed, one of which being the low accuracy w.r.t. clean labels in the presence of backdoor perturbations.

\subsection{Multi-Agent Backdoor Attacks}

After $N$ attackers contribute private datasets $D_i$ to joint dataset $\mathbb{D}$, the trained model evaluates backdoored inputs to compute the payoffs.
Let $\mathcal{X} \in \mathbb{R^{\textnormal{l} \times \textnormal{w} \times \textnormal{c}}}$, $\mathcal{Y} = {1, 2, . . . , k}$, $\Theta$ be the input, output, and parameter space respectively.
$\{D_i\}^{N}, \mathbb{D} \setminus \{D_i\} \sim \mathcal{X} \times \mathcal{Y}$
are sources of shifted $\mathcal{X}$:$\mathcal{Y}$ distributions from which an observation $x$ can be sampled.
$x$ can be decomposed $x = \mathbf{x} + \varepsilon$, where $\mathbf{x}$ is the set of clean features in $x$, and $\varepsilon : \{ \varepsilon \geq 0 \}^{N+1}$ is the set of perturbations that can exist.
$f$ is a base learner function
that accepts inputs $x$, $\theta$ to return predicted labels $\hat{y} = f(x; \theta)$. 
The model parameters $\theta \sim \Theta$ are sampled from the parameter space
such that it minimizes the loss between the ground-truth and predicted labels: 
$\mathcal{L}(\theta; x, y) = \frac{1}{|x|} \sum_{i}^{|x|} (f(x; \theta)-y)^2$.

Each attacker is a player $\{a_{i}\}^{i \in N}$ that 
generates backdoored inputs $X^{\textnormal{poison}}$ to insert into their private dataset contribution $\{X^{\textnormal{poison}} \in D_i\}^{i \in N} \in \mathbb{D}$. 
Attackers use backdoor attack algorithm $b_i$, which accepts a set of inputs mapped to target poisoned labels $\{X_i:Y_i^{\textnormal{poison}}\} \in D_i$ to specify the intended label classification, backdoor perturbation rate $\varepsilon_i$ to specify the proportion of an input to be perturbed, and the poison rate $p_i$ to specify the proportion of the private dataset to contain backdoored inputs, to return $X^{\textnormal{poison}} = b_i(X_i, Y_i^{\textnormal{poison}}, \varepsilon_i, p_i)$. 
An attacker $a_i$ aims to maximize the accuracy of the predicted outputs in test-time against the target poisoned labels as their payoff (Eqt \ref{equation:attacker_obj}), the \textit{attack success rate} (ASR), which is the rate of misclassification of backdoored inputs $X^{\textnormal{poison}}$, from the clean label $Y_i^{\textnormal{clean}}$ to the target poisoned label $Y_i^{\textnormal{poison}}$, by the defender's model $f$.
Each attacker optimizes their actions against the collective set of actions of the other $\neg i$ attackers.
\begin{equation}
\small
\setlength{\abovedisplayskip}{5pt}
\setlength{\belowdisplayskip}{5pt}
\begin{split}
x_i^{poison} & = x_i \odot (1-z_i) + m_i \odot z_i \\
b: X_i^{\textnormal{poison}} &:= \{ x_i \odot (1-z_i) + m_i \odot z_i \}^{x_i \in X_i^{\textnormal{poison}}}
\end{split}
\label{equation:poison}
\end{equation}
We adopt $b$: \textit{Random-BadNet} (Eqt \ref{equation:poison}) \cite{datta2022backdoors}, an extension of the baseline backdoor attack algorithm BadNet \citep{8685687} with the adaptation that, instead of a single square trigger pattern, we generate randomized pixels such that each attacker has their own unique trigger pattern. 
Within the given dimensions ($\texttt{length } l \times \texttt{width } w \times \texttt{channels } c$) of an input $x \in X$,
a backdoor trigger pattern is implemented as a mask $m_i$ that replaces pixel values of $x_i$.
$m_i$ is a randomly-generated trigger pattern, sampled per attacker $a_i$.
$z_i$ is the binary mask (corresponding to $m_i$) returning 1 at the location of a perturbation and 0 everywhere else.
$\odot$ is the element-wise product operator.
Perturbation rate $\varepsilon_i$ dictates the likelihood that an index pixel $(l, w, c)$ will be perturbed, and is used to generate the shape mask. 
A higher $\varepsilon_i$ results in higher density of perturbations. 
The poison rate is the proportion of the private dataset that is backdoored:
$p = \frac{|X^{\textnormal{poison}}|}{|X^{\textnormal{clean}}| + |X^{\textnormal{poison}}|}$.

A defender trains a model $f$ on the joint dataset $\mathbb{D}$, which may contain backdoored inputs, until it obtains model parameters $\theta$.
The defender can query the joint dataset and contributions $\mathbb{D}$; whether the index of the train-time agent is known at test-time is also evaluated in this work.
The defender can choose a model architecture (action $r_j$) and backdoor defense (action $s_j$).
The multi-agent backdoor attack is focused on minimizing the collective attack success rate, hence the defender's payoff can be approximated as the complementary of the mean attacker payoff (Eqt \ref{equation:defender_obj}).
\begin{numcases}{}
\begin{split}
\pi^{a_i} = 
\texttt{Acc}\Big(f(X_i; (\theta, \mathbb{D}); (r_j, s_j); \\ \{(\varepsilon_i, p_i, Y_i^{\textnormal{poison}}, b_i), \\ (\varepsilon_{\neg i}, p_{\neg i}, Y_{\neg i}^{\textnormal{poison}}, b_{\neg i}) \}), \\ Y_i^{\textnormal{poison}} \Big)  
\end{split}
\label{equation:attacker_obj}
\\
\begin{split}
\pi^{d} = 1- \frac{1}{N} \sum_i^{N} \texttt{Acc}(f(\cdot), Y_i^{\textnormal{poison}})
\end{split}
\label{equation:defender_obj}
\end{numcases}
We denote the collective attacker payoff and defender payoff as $\pi^{a} = \texttt{mean} \pm \texttt{std}$ and $\pi^{d} =((1-\texttt{mean}) \pm \texttt{std})$ respectively.
The equilibrium $\widetilde{\pi}^{a, d}$ is the collective payoff 
($\pi^{a}$, $\pi^{d}$) where both payoff values are maximized with respect to the dominant strategy taken by the other. 
Through the \textit{backfiring effect}, Datta \& Shadbolt (2022) \cite{datta2022backdoors} demonstrate the equilibrium $\widetilde{\pi}^{a, d}$ tends to $\frac{1}{|\mathcal{Y}|}$.

\subsection{Backfiring effect}

In a backdoor attack, the predicted label can be evaluated against the poison label or the clean label, resulting in 3 metrics:
\circled{1} run-time accuracy of the predicted labels with respect to (w.r.t.) poisoned labels given backdoored inputs,
\circled{2} run-time accuracy of the predicted labels w.r.t. clean labels given clean inputs,
\circled{3} run-time accuracy of the predicted labels w.r.t. clean labels given backdoored inputs.
The defender's primary objective is to minimize the individual and collective attack success rate of a set of attackers (minimize \circled{1}), and its secondary objective is to retain accuracy against clean inputs (maximize \circled{2}). 
In the traditional backdoor setting, the defender is less concerned with maximizing \circled{3} as long as \circled{1} is minimized.
The \textit{backfiring effect} has been beneficial as a natural defense against backdoor attacks, in particular with respect to the traditional metrics \circled{1} and \circled{2}. 

\begin{figure*}[!htbp]
    \centering
    \includegraphics[width=0.65\textwidth]{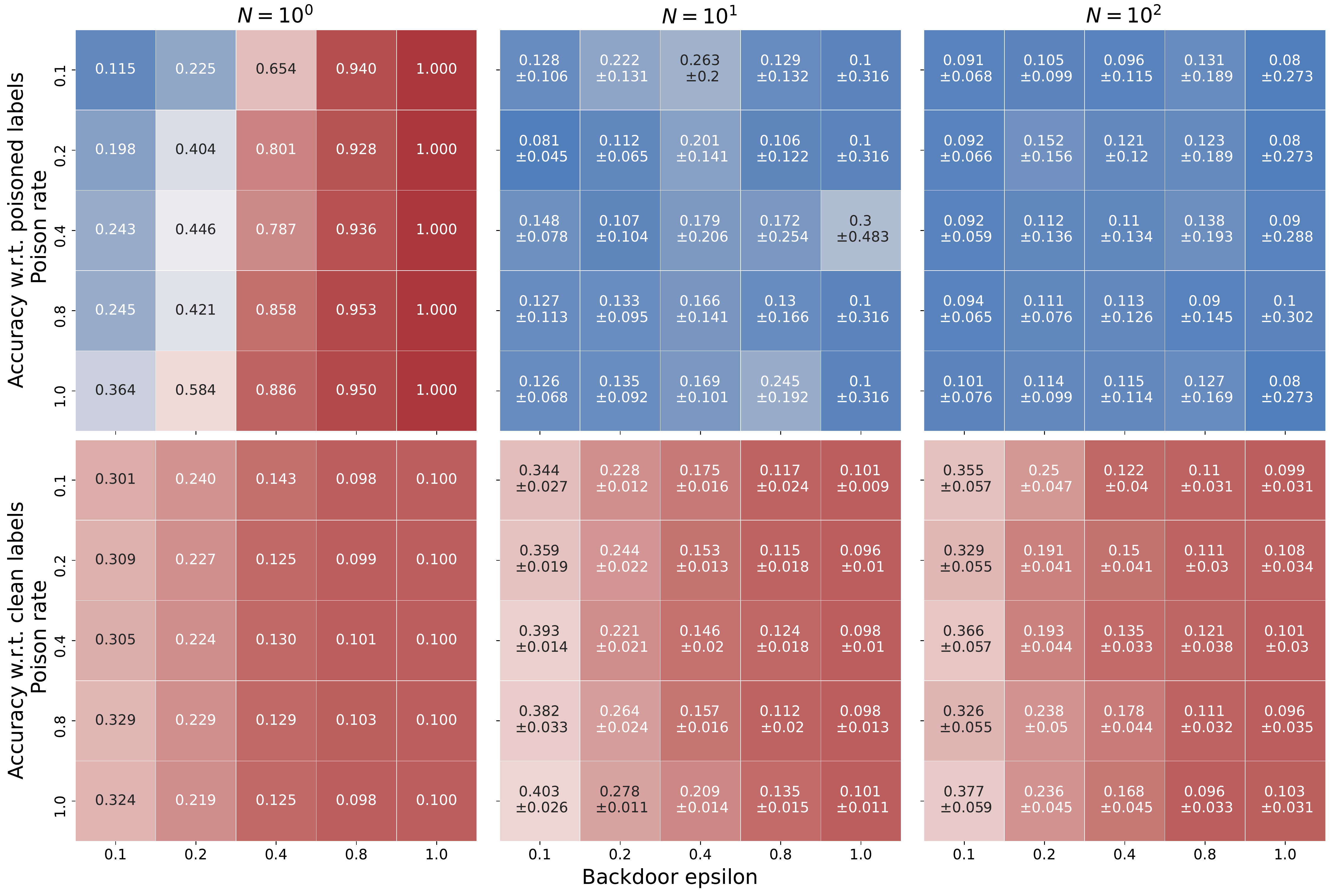}
    \caption{
    Side-effect from the \textit{backfiring effect}: Though mitigable in the single-agent setting, single-agent defenses cannot sufficiently mitigate the loss in accuracy w.r.t. clean labels in the multi-agent setting.
    }
    \label{fig:backfire}
\end{figure*}

\textbf{(Observation 0: Backfire results in reduced accuracy w.r.t. clean labels against backdoored inputs)}
An observation made in Datta \& Shadbolt (2022) \cite{datta2022backdoors}, denoted as the \textit{backfiring effect}, is that if $N \rightarrow \infty$, then the predicted label is uniformly sampled $y^{*} = f(\mathbf{x} + \varepsilon; \theta) \sim \mathcal{U}(\mathcal{Y})$ s.t. $\mathbf{P}(y^{*})=\frac{1}{|\mathcal{Y}|}$.
Refactoring this for an $\ell$-layer neural network, if the $(\ell-1)$th layer is the layer before the prediction layer that returns class probabilities, and $\phi_i$ is a specific subnetwork responsible for classifying $\{\varepsilon_i \mapsto \phi_i\}$, 
then it follows that, while the \textit{backfiring effect} induces $f_{\ell-1}(\varepsilon; \theta+\phi_{i}) = \{ y: \frac{1}{|\mathcal{Y}|} \}$, it also induces random sampling for clean labels. 
\begin{equation}
\small
\setlength{\abovedisplayskip}{5pt}
\setlength{\belowdisplayskip}{5pt}
\begin{split}
y^{*} & = f(\mathbf{x} + \varepsilon; \theta) \sim \mathcal{U}(\mathcal{Y}) \textnormal{ s.t.} \\
y^{*} & := \argmax_{y \sim \mathcal{Y}} \{ f_{\ell-1}(\mathbf{x}; \theta) + \sum_{i}^{N+1}f_{\ell-1}(\varepsilon; \theta+\phi_{i})\} \\
& := \argmax_{y \sim \mathcal{Y}} \{ f_{\ell-1}(\mathbf{x}; \theta) + \{ y: \frac{1}{|\mathcal{Y}|} \} \} \\
\end{split}
\label{equation:backfire}
\end{equation}
Validating with empirical evaluation (Figure \ref{fig:backfire}), 
we find that one of the side effects from the \textit{backfiring effect} is not merely, given a backdoored input that the accuracy w.r.t. poison labels tend to $\frac{1}{|\mathcal{Y}|}$, but the accuracy w.r.t. clean labels also tends to $\frac{1}{|\mathcal{Y}|}$ .
Hence, to mitigate \circled{3}, we explore methods that result in $f_{\ell-1}(\varepsilon; \theta+\phi_{i}) = \{ y: 0 \}$, i.e. $|\phi_i| \rightarrow 0$.

The question that motivates this paper is what strategy can a defender take to further maximize its welfare? To suppress the equilibrium accuracy w.r.t. poison labels, it is shown in Datta \& Shadbolt (2022) \cite{datta2022backdoors} that a (single-agent) defense should be employed. Since the accuracy w.r.t. poison labels is already at the lower bound, it is not meaningful to pursue a variant multi-agent defense that further attempts to suppress the accuracy w.r.t. poison labels, at least given the current knowledge of its failure modes. 
Hence, to make the most out of a deployed defense and maximize defender welfare, we motivate this work on multi-agent defenses to jointly optimize all 3 metrics, in particular maximizing accuracy w.r.t. clean labels in the presence of backdoor perturbations.

\subsection{Compressed Low-Loss Subspaces}

The \textit{parameter space} $\Theta \in \mathbb{R}^M$ is a topological space, where $M$-dimensional parameter point-estimates $\theta \sim \Theta$ are sampled and loaded into a base learner function $f(\theta; \cdot)$.
A \textit{parameter subspace} $\vartheta$ is a bounded subspace contained in $\Theta$ such that: 
\begin{equation}
\footnotesize
\begin{split}
\vartheta = \Big\{ \Sigma_i^N \alpha_i \theta_i \in \Theta \Big| & 0 \leq \alpha_i \leq 1, \\
& 0 \leq \Sigma_i^N \alpha_i \leq N, \\
& \{ \min \mathcal{L}(\theta_i; X_i, Y_i) \}^{i \in N} \Big\}
\end{split}
\end{equation}
A \textit{compressed low-loss subspace} is a parameter subspace with the added constraint that the distance between the set of end-point parameters 
should be minimized (Algorithm \ref{alg:train}) such that:
\begin{equation}
\footnotesize
\begin{split}
\vartheta = \Big\{ \Sigma_i^N \alpha_i \theta_i \in \Theta \Big| & 0 \leq \alpha_i \leq 1, \\
& 0 \leq \Sigma_i^N \alpha_i \leq N, \\
& \{ \min \mathcal{L}(\theta_i; X_i, Y_i) \}^{i \in N}, \\ 
& \min \textnormal{\texttt{dist}}(\{\theta_i\}^N)
\Big\}
\end{split}
\end{equation}

\noindent\textbf{Theorem 1. }
\textit{
Evaluating a backdoored input $x_i = \mathbf{x} + \varepsilon_i$ against an interpolated parameter $\theta_{\widehat{ij}}$ sampled from a compressed low-loss subspace will tend to return a higher accuracy w.r.t. clean labels.
}

\noindent\textbf{Proof sketch of Theorem 1. }
Suppose $N$ parameters are trained in parallel to construct a compressed low-loss subspace, where each parameter is mapped to a shifted distribution $\mathbf{x} + \varepsilon_{i} \mapsto \theta_i$.
For simplicity, we assume $f(x; \theta) = \theta x$.
We sample any two end-point parameters, $\theta_i$ and $\theta_j$, and linearly-interpolate along this compressed low-loss line.

We denote the following loss values: \newline$\mathcal{L}_1 = \mathcal{L}(\theta_i; x_i, y_i^{\textnormal{poison}})$,
$\mathcal{L}_2 = \mathcal{L}(\theta_i; x_i, y_i^{\textnormal{clean}})$,
\newline$\mathcal{L}_3 = \mathcal{L}(\theta_j; x_i, y_i^{\textnormal{poison}})$,
$\mathcal{L}_4 = \mathcal{L}(\theta_j; x_i, y_i^{\textnormal{clean}})$.
Based on the definition of backdoor attack,
we know that $\mathcal{L}_1 < \mathcal{L}_2, \mathcal{L}_3$; $\mathcal{L}_4 < \mathcal{L}_2$; $\mathcal{L}_4 \approx \mathcal{L}_1$.
$\mathcal{L}_3 \rightarrow \infty$ (i.e. rejected). 

First, we compute the loss of $\theta_i$ w.r.t. $x_i$ under distance-regularization (squared Euclidean distance). 
\begin{equation}
\footnotesize
\begin{split}
\mathcal{L}(\theta_i) &= (\theta_i x_i - y_i^{\textnormal{poison}}) + (\theta_i - \theta_j)^2\\
&= \theta_i x_i - y_i^{\textnormal{poison}} + \theta_i^2 - 2\theta_i \theta_j + \theta_j^2 \\
&= \theta_i^2 - \theta_i (x_i - 2\theta_j) + (\theta_j^2 - y_i^{\textnormal{poison}})
\end{split}
\end{equation}
Upon convergence of all loss terms $\mathcal{L}(\theta_i) \rightarrow 0$, 
\begin{equation}
\footnotesize
\begin{split}
\theta_i &:= \frac{2\theta_j-x_i \pm \sqrt{(x_i - 2 \theta_j)^2 - 4(\theta_j^2-y_i^{\textnormal{poison}})}}{2} \\
\theta_i &:= \frac{2\theta_j-x_i \pm \sqrt{x_i^2 - 2 x_i \theta_j + 4 y_i^{\textnormal{poison}}}}{2}
\end{split}
\label{equation:bk1}
\end{equation}
And vice versa, 
\begin{equation}
\footnotesize
\begin{split}
\theta_j &:= \frac{2\theta_i-x_j \pm \sqrt{x_j^2 - 2 x_j \theta_i + 4 y_j^{\textnormal{poison}}}}{2}
\end{split}
\label{equation:bk2}
\end{equation}
Eqts \ref{equation:bk1} and \ref{equation:bk2} show that each boundary parameter is a function of the opposing parameter(s) in addition to their mapped datasets. 
Subsequently, we can compose an interpolated parameter $\theta_{\widehat{ij}}$ weighted by interpolation coefficient $\alpha \sim [0, 1]$ as follows.
\begin{equation}
\footnotesize
\begin{split}
\theta_{\widehat{ij}} & = \alpha \theta_i + (1-\alpha) \theta_j \\
&= 
\frac{\alpha}{2} 
\bigg[ 
2 \theta_j - x_i \pm \sqrt{x_i^2 - 2x_i \theta_j + 4y_i^{\textnormal{poison}} }
\bigg] \\
& \phantom{=} 
+ \frac{1-\alpha}{2} 
\bigg[ 
2 \theta_i - x_j \pm \sqrt{x_j^2 - 2x_j \theta_i + 4y_j^{\textnormal{poison}} }
\bigg] \\
&=
2 \alpha \theta_j - \alpha x_i + 2 \theta_i - 2 \alpha \theta_i - x_j + \alpha x_j \\
& \phantom{=}
\pm \alpha \sqrt{x_i^2 - 2 x_i \theta_j + 4y_i^{\textnormal{poison}}} \\
& \phantom{=}
\frac{\pm (1-\alpha) \sqrt{x_j^2 - 2 x_j \theta_i + 4y_j^{\textnormal{poison}}}
}{2}
\end{split}
\end{equation}
Next, we compute the predicted output $\theta_{\widehat{ij}}$ w.r.t. $x_i$ to observe how loss varies.
\begin{equation}
\footnotesize
\begin{split}
\theta_{\widehat{ij}} x_i
&=
2 \alpha \theta_j x_i - \alpha x_i^2 + 2 \theta_i x_i - 2 \alpha \theta_i x_i + (1-\alpha) x_i x_j \\
& \phantom{=}
\pm \alpha \sqrt{x_i^4 - 2 x_i^3 \theta_j + 4y_i^{\textnormal{poison}}} \\
& \phantom{=}
\frac{\pm (1-\alpha) \sqrt{x_i^2 x_j^2 - 2 x_i^2 x_j \theta_i + 4 x_i^2 y_j^{\textnormal{poison}}}
}{2} \\
&=
2 \alpha (\mathcal{L}_4 + y_i^{\textnormal{clean}}) - \alpha x_i^2 + 2 (1- \alpha) (\mathcal{L}_1 + y_i^{\textnormal{poison}}) \\
& \phantom{=}  + (1-\alpha) x_i x_j
\pm \alpha \sqrt{x_i^4 - 2 x_i^2 (\mathcal{L}_4 + y_i^{\textnormal{clean}}) + 4y_i^{\textnormal{poison}}} \\
& \phantom{=}
\frac{\pm (1-\alpha) \sqrt{x_i^2 x_j^2 - 2 x_i x_j (\mathcal{L}_1 + y_i^{\textnormal{poison}}) + 4 x_i^2 y_j^{\textnormal{poison}}}
}{2}
\end{split}
\end{equation}

Hence, 
with respect to $x_i$, 
when $\alpha \rightarrow 1$,
then $\mathcal{L}_4$ manifests to a larger magnitude than $\mathcal{L}_1$ such that
$\theta_{\widehat{ij}} x_i - y_i^{\textnormal{clean}} < \theta_{\widehat{ij}} x_i - y_i^{\textnormal{poison}}$.
Therefore, 
we observe that the accuracy w.r.t. clean labels should increase when interpolating within a parameter subspace regularized by distance.

\qed

%
\vfill
\begin{algorithm}[b]
  \footnotesize 
  \caption{AgentSubspace: Train}
  \SetKwInOut{Input}{Input}
  \SetKwInOut{Output}{Output}
  \SetKwProg{trainAgentSubspace}{trainSpace}{}{}
  \trainAgentSubspace{$(f, \{\theta_i\}^{N}_{i=1}, \{D_i\}^{N}_{i=1}, \beta, E$)}{
    \Input{model $f$, parameters $\{\theta_i\}^{N}_{i=1}$, train set $\{D_i\}^{N}_{i=1}$,  coefficient $\beta$, epochs $E$}
    \Output{Trained parameters $\{\theta_i\}^{N}_{i=1}$}
    Constant initialization $\{\theta_{\textnormal{init}} \gets \theta_i\}^{N}_{i=1}$
    \smallbreak
    Agent-indexed sets $\{S_i\}^{N}_{i=1} \gets \{ \bigcup_{j \neq i}^{N} D_j \}^{N}_{i=1}$ \\
    Train each parameter against their agent-indexed set in parallel per epoch \\
    \ForEach{\textnormal{epoch} $e \in E$}{
        \ForEach{$\theta_i, S_i \in  \{\theta_i\}^{N}_{i=1}, \{S_i\}^{N}_{i=1}$}{
            \ForEach{\textnormal{batch} $x, y \in S_i$}{
                $\mathcal{L} \gets 
                \frac{1}{|x|} \sum_{i}^{|x|} (f(\theta_i; x)-y)^2$ \\
                \phantom{$\mathcal{L} \gets$} $+ 
                \frac{\beta}{N-1} \sum_{j=1, j \neq i}^{N-1} \textnormal{cos}(\theta_i, \theta_j)
                $\\
                Backprop $\theta_i$ w.r.t. $\mathcal{L}$
            }
        }
    }
    \KwRet{$\{\theta_i\}^{N}_{i=1}$}\; 
  }
  \label{alg:train}
\end{algorithm}

\begin{algorithm}[b]
  \footnotesize 
  \caption{AgentSubspace: Evaluate}
  \SetKwInOut{Input}{Input}
  \SetKwInOut{Output}{Output}
  \SetKwProg{evalSpace}{evalSpace}{}{}

    \evalSpace{$x, \{\theta_i\}_{i=1}^{N}, \{\alpha_i\}_{i=1}^{N}$}{
    \Input{Input $x$, Trained parameters $\{\theta_i\}_{i=1}^{N}$, Interpolation coefficients $\alpha_i \sim [0, 1]$ s.t. $\sum_{i=1}^{N} \alpha_i \leq N$}
    \Output{Predicted output $\hat{y}$}
    $\theta \gets \sum_{i=1}^{N} \alpha_i \theta_i$\\
    $\hat{y} \gets f(x, \theta)$\\
    \KwRet{$\hat{y}$}\;
  }
  \label{alg:inf}
\end{algorithm}

Constructing subspaces enables us to sample diverse ensembles and parameters, which has been indicated to perform well against distribution shifts.
Fort et al. (2020) \cite{fort2020deep} and Wortsman et al. (2021) \cite{wortsman2021learning}
validate the robustness of sampled ensembles against adversarial perturbations.
Wortsman et al. (2021) \cite{wortsman2021learning} validates its ability to improve general uncertainty estimation.
Datta \& Shadbolt (2022) \cite{https://doi.org/10.48550/arxiv.2205.09891}
construct a compressed parameter subspace to 
return optimal parameters for various test-time distribution shifts,
including label shift (backdoor, adversarial, permutation, rotation perturbations),
domain shift, and task shift.
Existing literature \citep{draxler2019essentially, garipov2018loss, frankle2020linear, fort2019large, goodfellow2015qualitatively, li2018measuring, izmailov2019subspace, izmailov2019averaging,  fort2020deep, wortsman2020supermasks, ratzlaff2020hypergan, vonoswald2021neural, wortsman2021learning, nunez2021lcs} investigate how to connect optima given random initializations.
We start with a constant initialization, and we would be amongst the first to evaluate the compression of a parameter subspace with respect to multiple different input:output distributions.
Prior work train for explicit pathways and connections between optima; we focus on compressing the space without explicit linear interpolation regularization, and find that random sampling points for ensembling can return sufficient low-loss parameters for inference, and evaluate its fitness as a robustness measure against joint distribution shift.

\section{Multi-Agent Backdoor Defenses}
\label{3}

\begin{figure*}
    \centering
    \includegraphics[width=\textwidth]{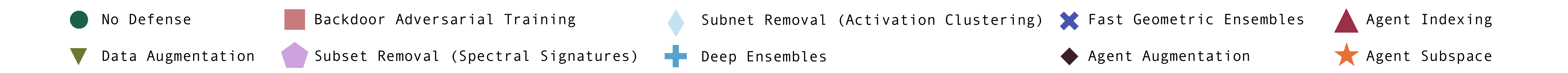}
    \begin{subfigure}{0.225\linewidth}
        \includegraphics[width=\linewidth]{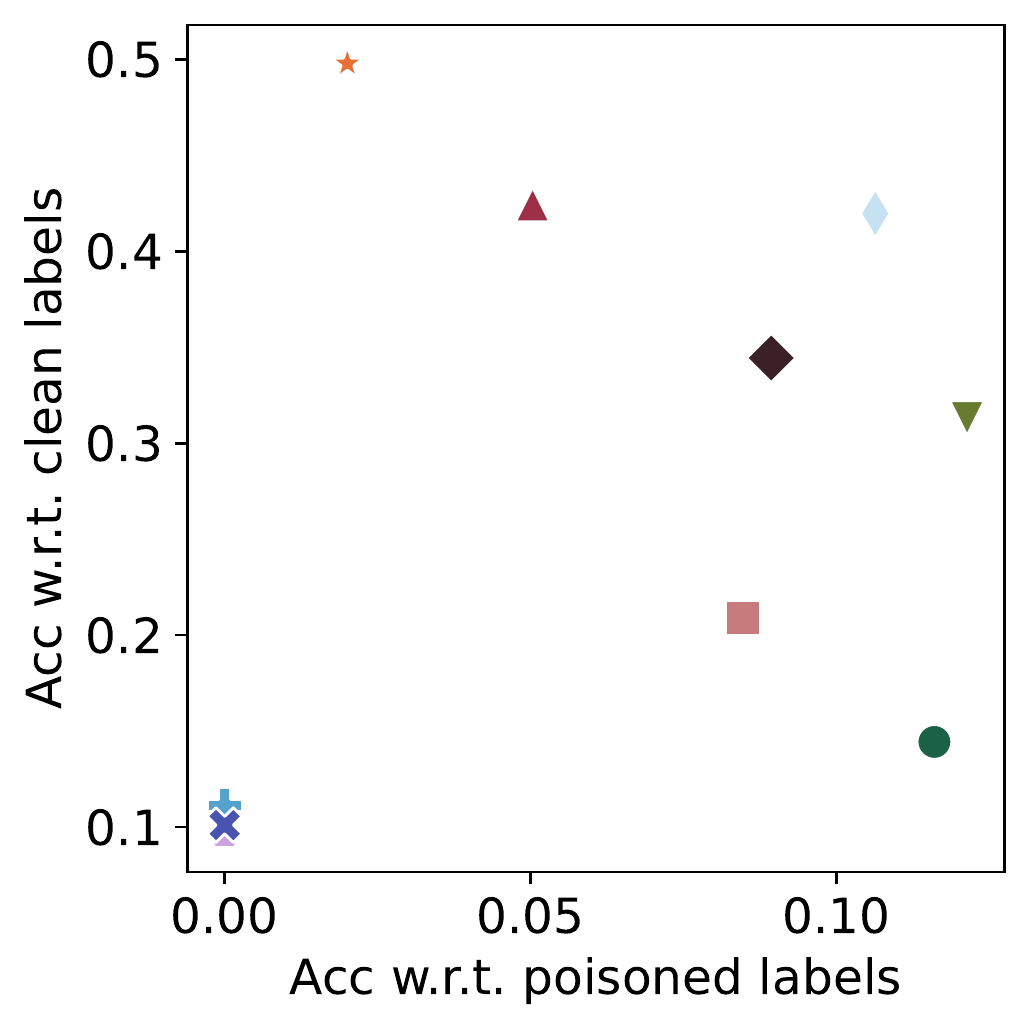}
        \caption{Base: CIFAR10, CNN, $\alpha=0$}
    \end{subfigure}
    \hfill
    \begin{subfigure}{0.225\linewidth}
        \includegraphics[width=\linewidth]{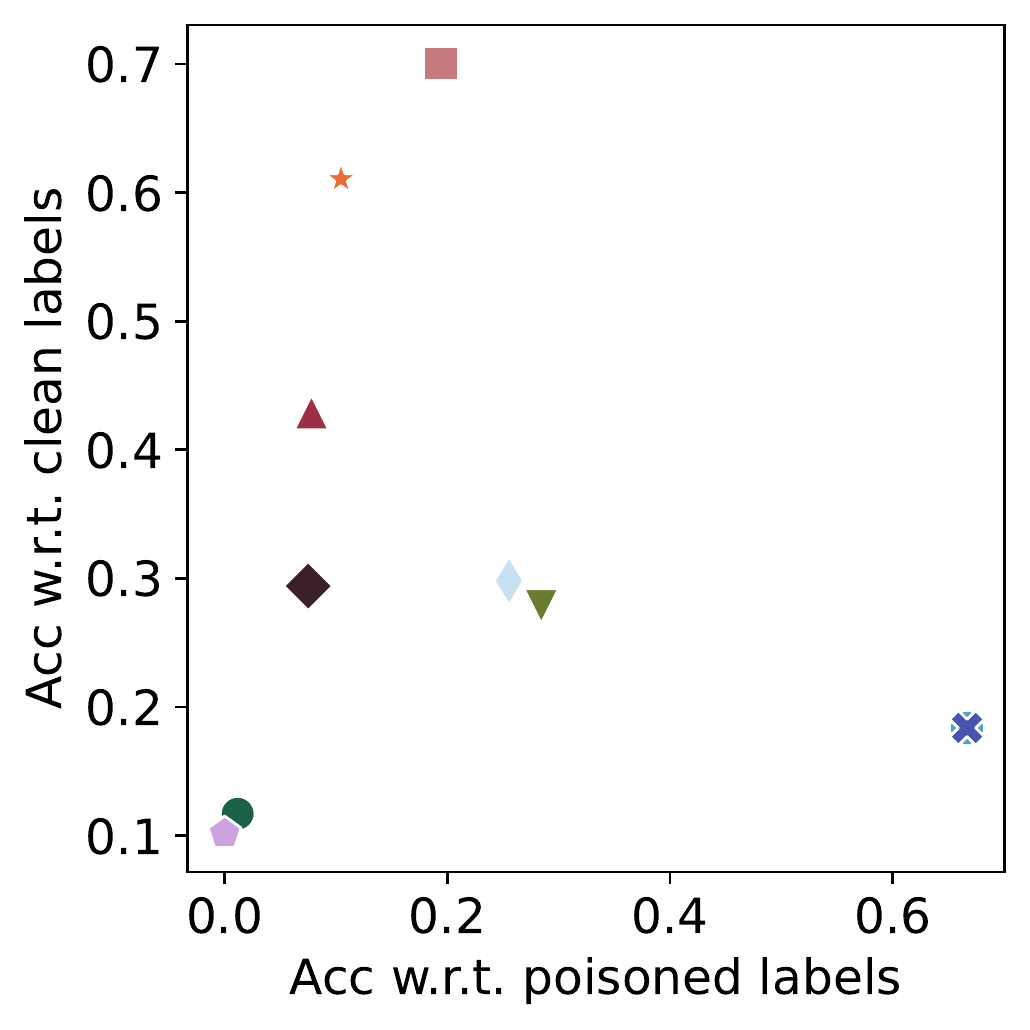}
        \caption{SVNH}
    \end{subfigure}
    \hfill
    \begin{subfigure}{0.225\linewidth}
        \includegraphics[width=\linewidth]{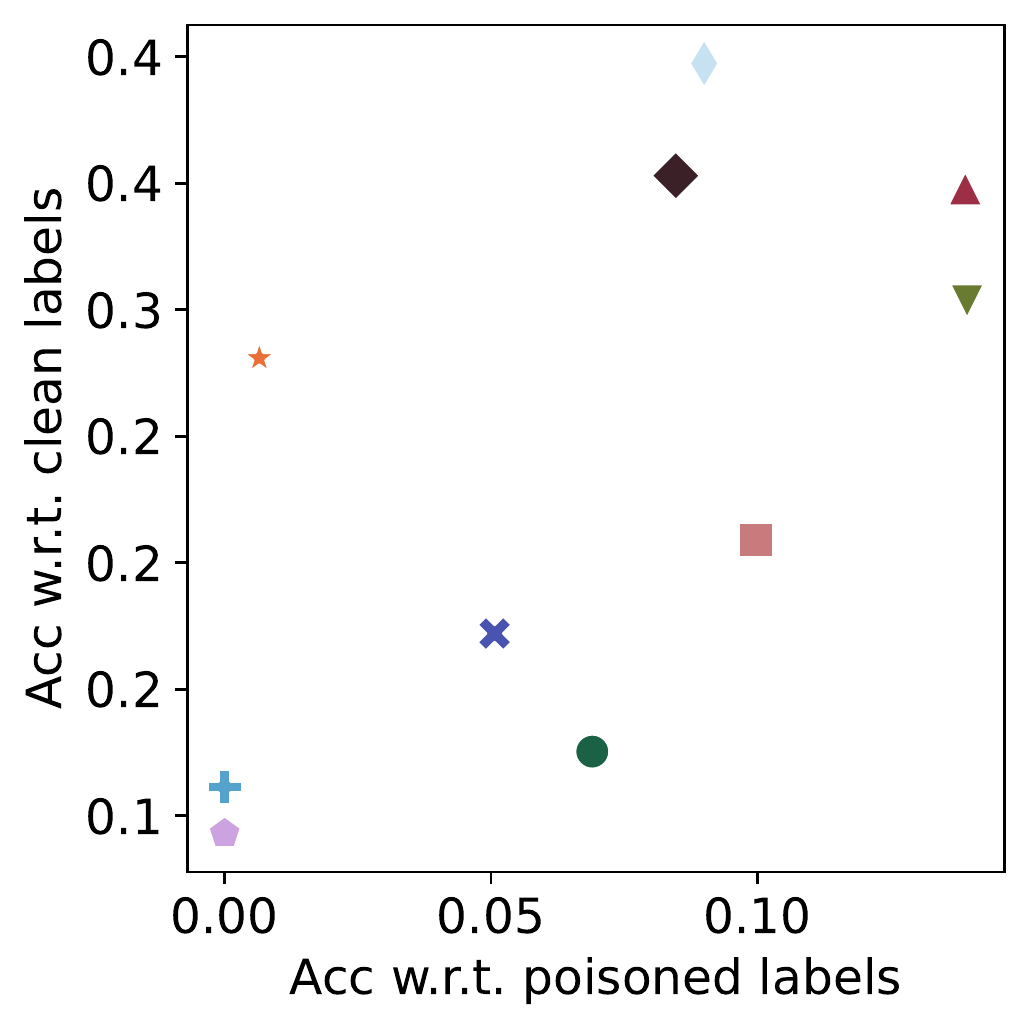}
        \caption{ResNet50}
    \end{subfigure}
    \hfill
    \begin{subfigure}{0.225\linewidth}
        \includegraphics[width=\linewidth]{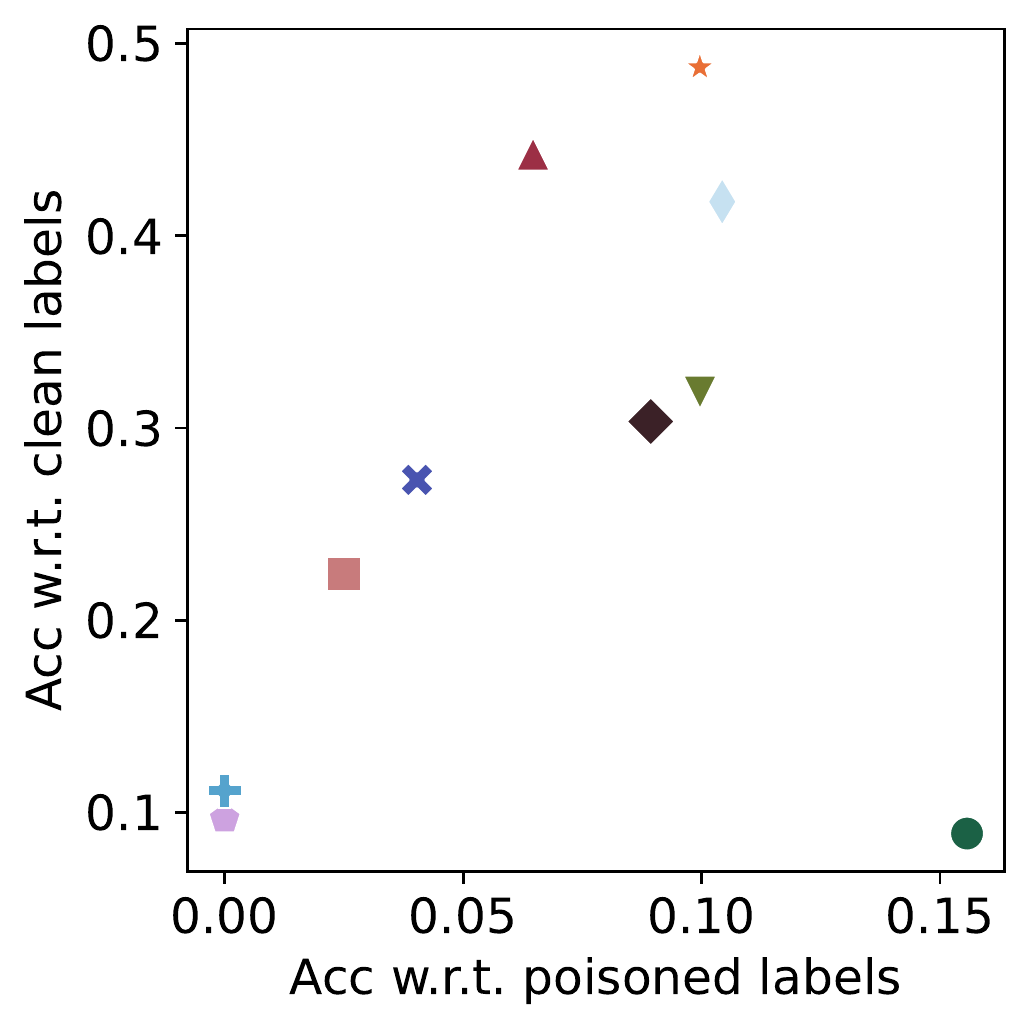}
        \caption{style $\alpha=1$}
    \end{subfigure}
  \caption{Trade-off between the accuracy w.r.t. poison and clean labels: The ideal backdoor defense should retain a high accuracy w.r.t. clean label (baseline: No Defense), while minimizing accuracy w.r.t. poison label. Ideal defenses are in the top-left region of each ablation plot.}
  \label{fig:ablations}
\end{figure*}

\subsection{Agent Subspace}

Based on our discussion in Section 2.3, we propose the Agent Subspace defense against multi-agent backdoor attacks, in particular to maximize the accuracy w.r.t. clean labels. Referring to Algorithm 1, 
we initialize $N$ parameters at the same point, train parameters in parallel such that the cosine distance between their weights matrices is minimized, where each parameter $\theta_i$ is evaluated on its corresponding agent indexing dataset.
During inference (Algorithm 2), we can compute with the subspace centre, or sample and ensemble $N=1000$ point-estimates.
We train the subspace for $5000$ epochs, with 3 attackers, and $\beta=0.05$.

\subsection{Agent Indexing}

In agent indexing, a baseline multi-agent backdoor defense, the defender has knowledge of agent index/identity between train-time and test-time
to isolate the learning effects introduced by data provided by specific agents (including triggers).
The intuition of this defense is to leverage the non-uniformity of backdoor perturbations and subnetworks between agents, as the backdoor subnetwork of $\neg i$ backfiring agents is unlikely to perform well across other unique backdoor perturbations.
In this case, we assume that the number of agents that interact in the system is known and that at test-time the defender knows which agent provided the sample to run inference on.
This form of inference, where an adaptation method changes parameters given some distributional meta-data or headers, is frequently used in other distribution shift settings such as continual learning or domain adaptation, and is also practically-relevant.

Rather than training a single model with all the available data, the defender trains a set of models, one for each agent in the system.
For an agent $a_i$, its corresponding model will only be trained on data $S_i$ provided by different agents $\neg i$, i.e. $\{S_i\}^{N}_{i=1} \gets \{ \bigcup_{j \neq i}^{N} D_j \}^{N}_{i=1}$.
At test-time, the defender selects which model to use based on the agent requesting the inference.

\subsection{Agent Augmentation}

In agent augmentation, a baseline multi-agent backdoor defense, a defender simulates the presence of fake backdoor attackers and  introduce simulated backdoored samples in the training data.
The implementation is similar to backdoor adversarial training~\citep{geiping2021doesnt} in the respect that we make use of the backdoor attack algorithm to poison our own dataset; the difference is that in order to simulate an augmented number of agents, we specifically set the number of agents to be high ($N=200$ here, while $N=20$ in backdoor adversarial training). 
This defense scales the \textit{backfiring effect} to ensure the accuracy w.r.t. poison labels stays low, but expectedly the accuracy w.r.t. clean labels should not be greatly improved.

Defenders can insert backdoor triggers into subsets of their clean dataset (the sub-dataset that the defender contributes). 
To evaluate agent augmentation, we allocate 40\% of the defender's contributed sub-dataset to be clean, and the remaining 60\% of samples to be poisoned with defender-generated triggers.
The defender sets the number of simulated attackers $n$, then generates unique random trigger patterns (one per attacker).
Each of the $n$ simulated attackers is allocated an equal amount of samples out of the total 60\%; we set the poison rate of simulated attackers $\varepsilon, p=0.2$.


\section{Evaluation}
\label{4}

\subsection{Methodology}

We evaluate upon CIFAR10 (10 classes, 60,0000 inputs, 3 colour channels) \citep{krizhevsky2009learning}.
The traintime-testtime split of each attacker is 80-20\% (80\% of the attacker's private dataset is contributed to the joint dataset, 20\% reserved for evaluating in test-time). 
The train-validation split for the defender was 80-20\% (80\% of joint dataset used for training, 20\% for validation). 
We trained a small CNN (channels $[16,32,32]$; $15,722$ parameters) model with batch size 128 with early stopping when loss converges (approximately $30$ epochs, validation accuracy of $92-93\%$; loss convergence depends on joint dataset structure and number of attackers).
We use a Stochastic Gradient Descent optimizer with $0.001$ learning rate and $0.9$ momentum, and cross entropy loss function.
For $N$ attackers and $V_d$ ($= 0.1$) being the proportion of the dataset allocated to the defender, 
the real poison rate (the absolute proportion of the joint dataset that is backdoored by attacker $a_i$) is calculated as  $\rho = (1 - V_d) \times \frac{1}{N} \times p$.
For attackers, we generate backdoor perturbations with Random-BadNet \cite{datta2022backdoors}, hold $p, \varepsilon = 0.4$, backdoor target labels are randomly chosen by each attacker.
Seed is 3407 for all procedures except those requiring each attacker to have distinctly different randomly-sampled values (e.g. trigger pattern generation), in which cases the seed is the index of the attacker. 

The primary variables 
are the existing single-agent and hypothesized multi-agent defenses.
We evaluate 2 augmentative (data augmentation, backdoor adversarial training), 2 removal (spectral signatures, activation clustering) defenses, and 2 ensembling methods (as baseline to our hypotheses). 
For augmentative defenses, $50\%$ of the defender's allocation of the dataset is assigned to augmentation: for $V_d = 0.8$, 0.4 is clean, 0.4 is augmented.
The number of trained point-estimates is set as the number of attackers, e.g. 3 ensemble models, 3 parameters trained in parallel for Agent Subspace.

\textbf{Data Augmentation~\citep{9414862}}
(e.g.,  CutMix~\citep{Yun_2019_ICCV} or MixUp \citep{zhang2018mixup}) has been shown to reduce backdoor attack success rate.
We implement CutMix~\citep{Yun_2019_ICCV}, where augmentation takes place per batch, and training completes in accordance with aforementioned early stopping.
%
\textbf{Backdoor Adversarial Training \cite{geiping2021doesnt}}
extend the concept of adversarial training on defender-generated backdoor examples to insert their own triggers to existing labels.
We implement backdoor adversarial training~\citep{geiping2021doesnt}, where the generation of backdoor perturbations is through Random-BadNet~\citep{datta2022backdoors}, 
where $50\%$ of the defender's allocation of the dataset is assigned to backdoor perturbation,
$p, \varepsilon = 0.4$, 
and 20 different backdoor triggers used
(i.e. allocation of defender's dataset per backdoor trigger pattern is
$(1-0.5) \times 0.8 \times \frac{1}{20}$). 

\textbf{Spectral Signatures~\citep{NEURIPS2018_280cf18b}}
is an input inspection method used to perform subset removal from a training dataset. 
For each class in the backdoored dataset, the method uses the singular value decomposition of the covariance matix of the learned representation for each input in a class in order to compute an outlier score, and remove the top scores before re-training.
In-line with existing implementations, we remove the top 5 scores for $N=1$ attackers. For $N=10$, we scale this value accordingly and remove the top 50 scores.
\textbf{Activation Clustering~\citep{chen2018detecting}}
is an input inspection method used to perform subset removal from a training set. 
In-line with Chen et al. (2018) \cite{chen2018detecting}'s implementation, we perform dimensionality reduction using Independent Component Analysis (ICA) on the dataset activations, then use \textit{k}-means clustering to separate the activations into two clusters, then use Exclusionary Reclassification to score whether a given cluster corresponds to backdoored data and remove it.

\textbf{Deep Ensembles \cite{lakshminarayanan2017simple}}
is a strong ensembling baseline for uncertainty estimation and sampling parameters for ensembling. It is constructed through training multiple networks independently in parallel with adversarial training. 
\textbf{Fast Geometric Ensembles \cite{garipov2018loss}}
train multiple networks to find their respective modes such that a high-accuracy path is formed that connects the modes in the parameter space. The method collects parameters at different checkpoints, which can subsequently be used as an ensemble.

\begin{figure*}[!htbp]
    \centering
    \includegraphics[width=0.225\textwidth]{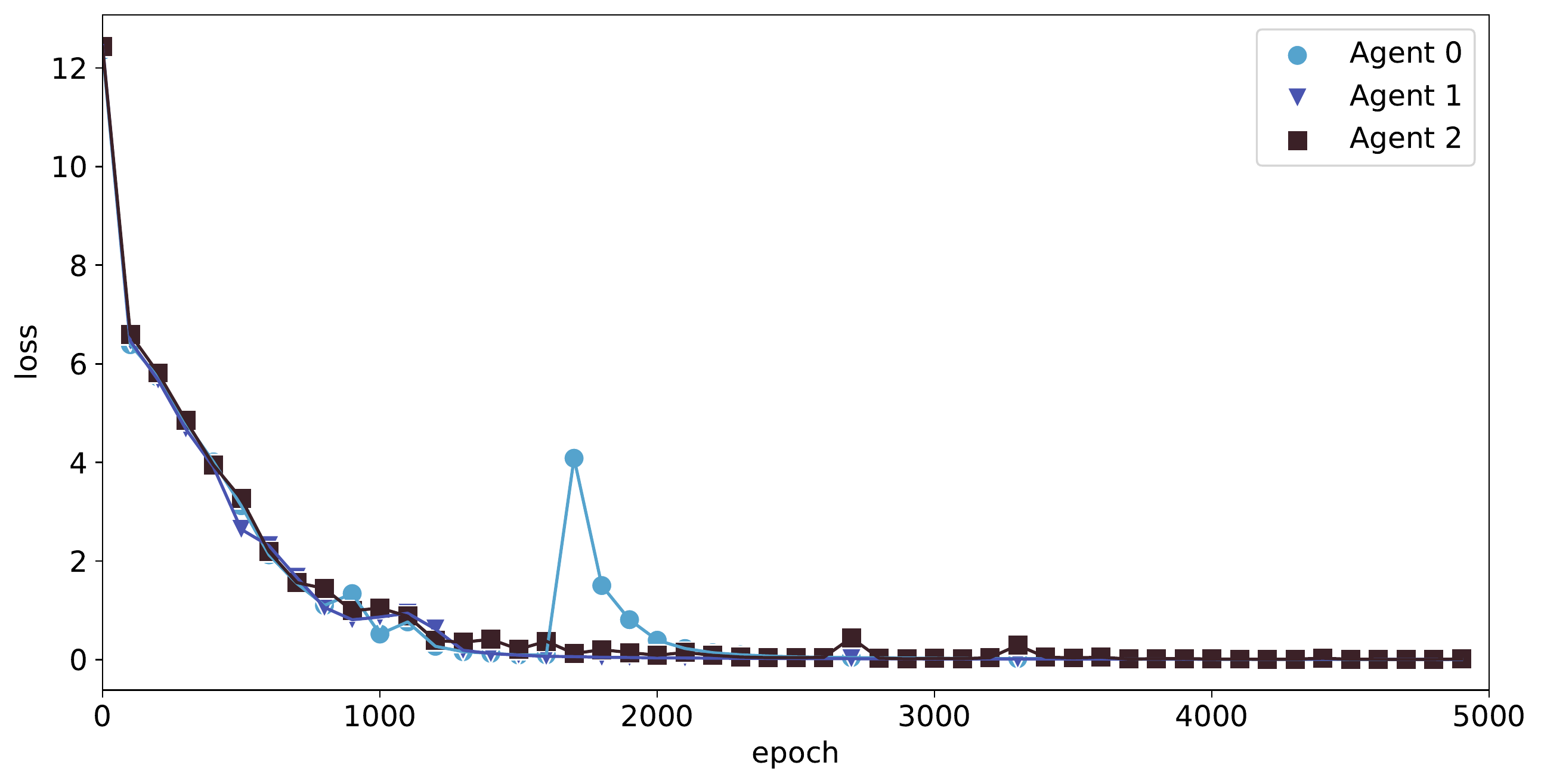}
    \includegraphics[width=0.225\textwidth]{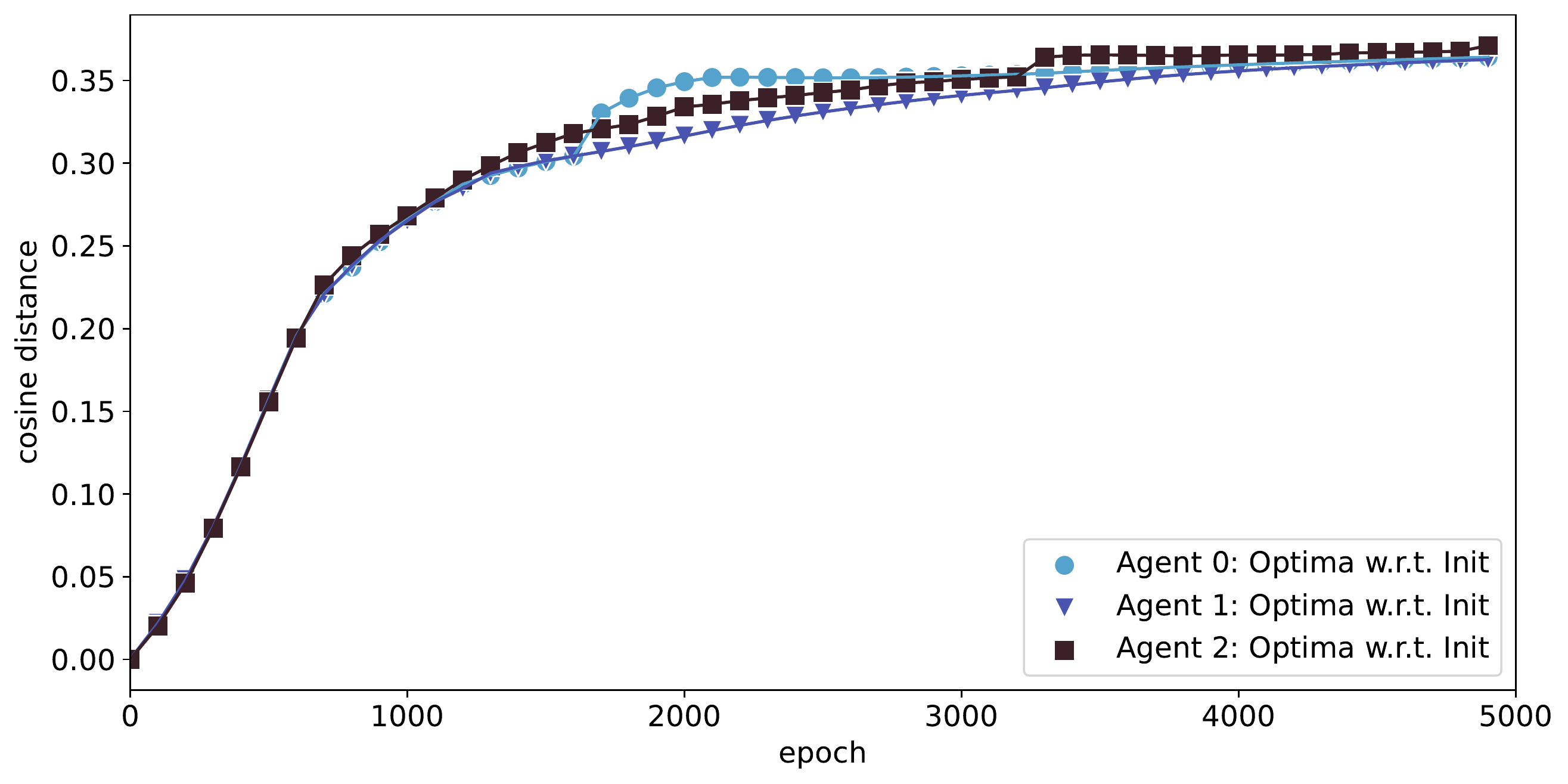}
    \includegraphics[width=0.225\textwidth]{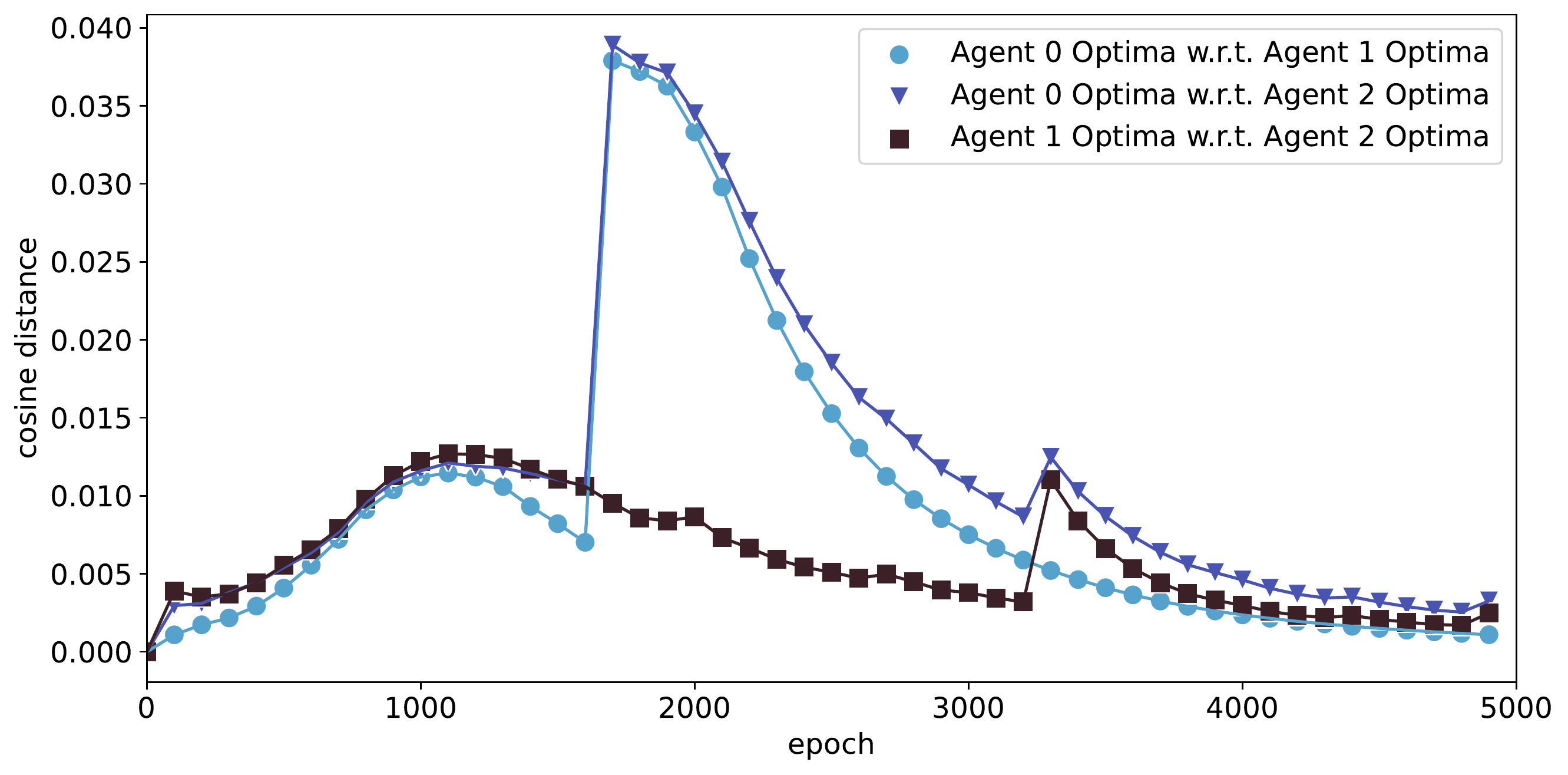}
    \includegraphics[width=0.225\textwidth]{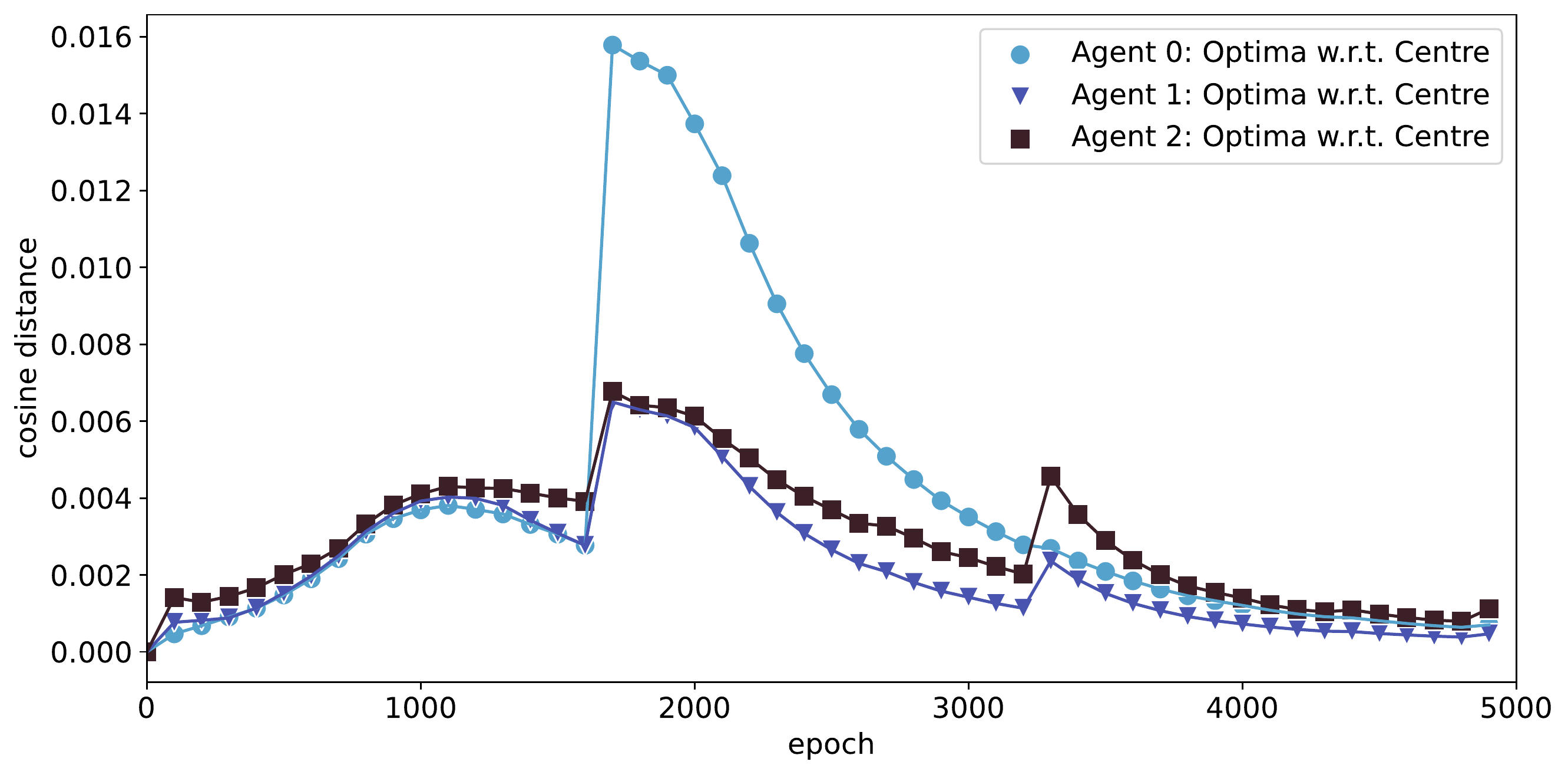}
    \caption{
    Convergence of loss w.r.t. respective inputs and cosine distance: as epoch increases, SGD finds an equilibrium subspace for sampling across distributions.
    }
    \label{fig:multi}
\end{figure*}
\begin{figure*}[h]
    \centering
    \includegraphics[width=0.225\textwidth]{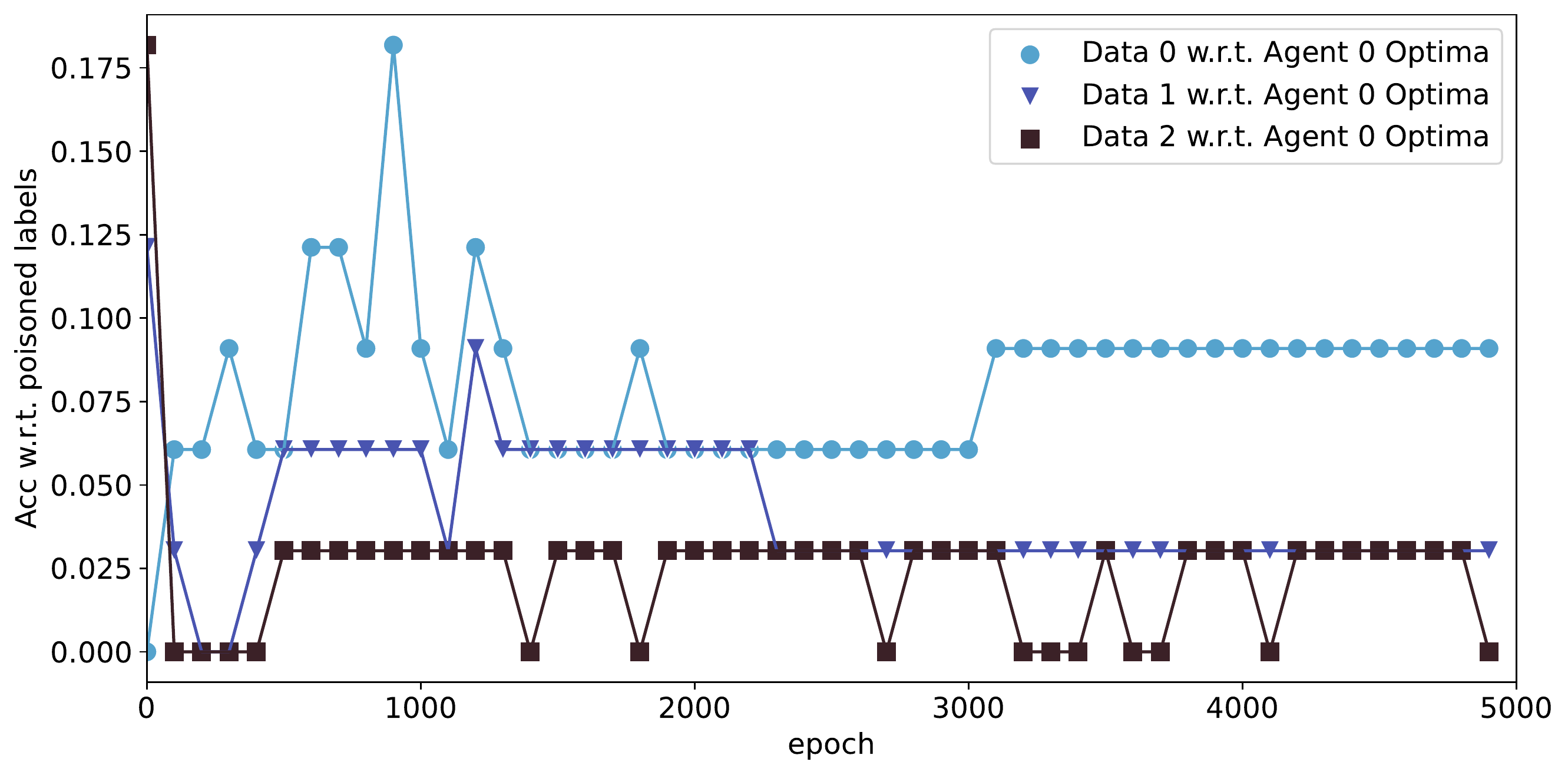}
    \includegraphics[width=0.225\textwidth]{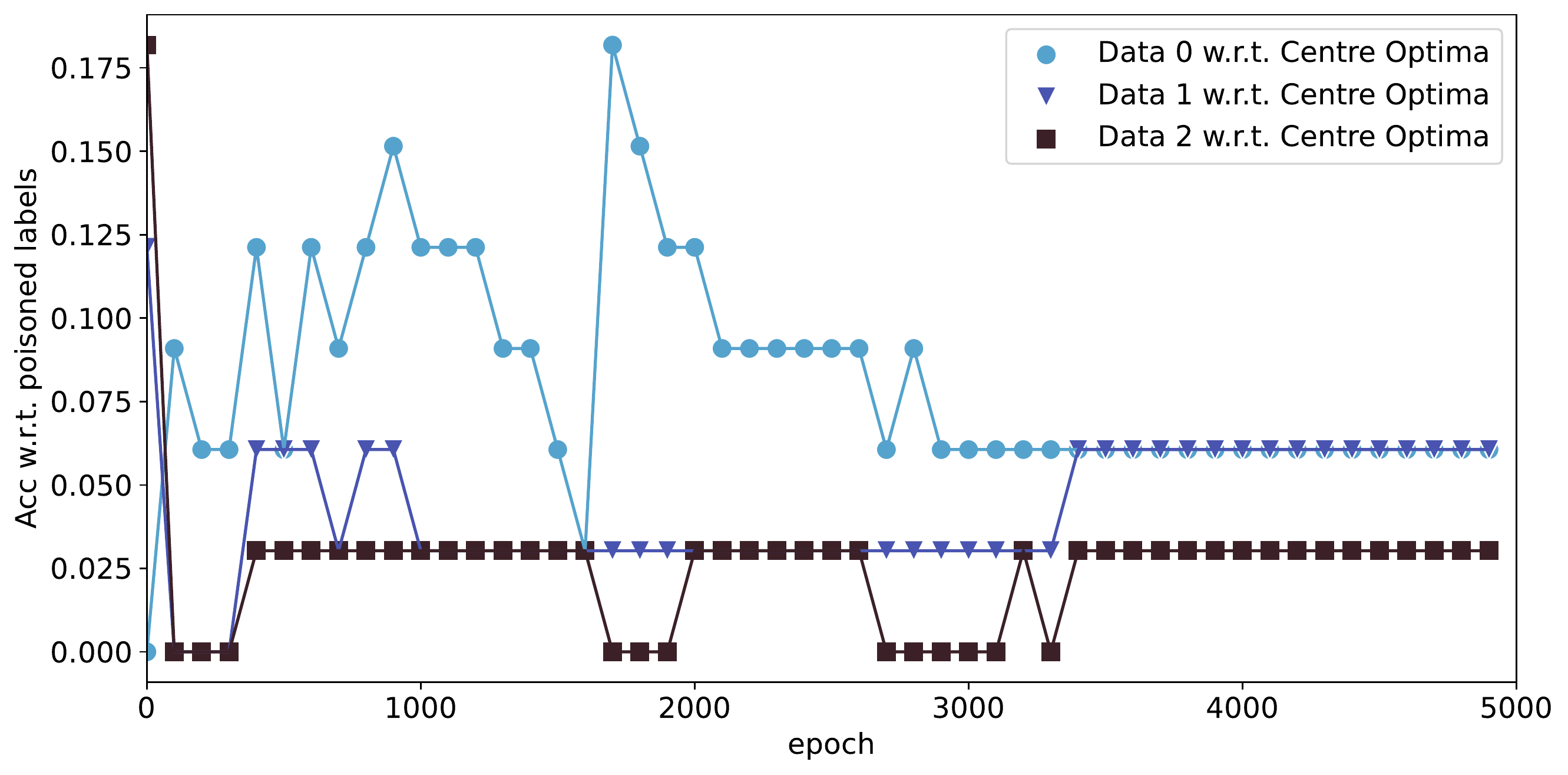}
    \includegraphics[width=0.225\textwidth]{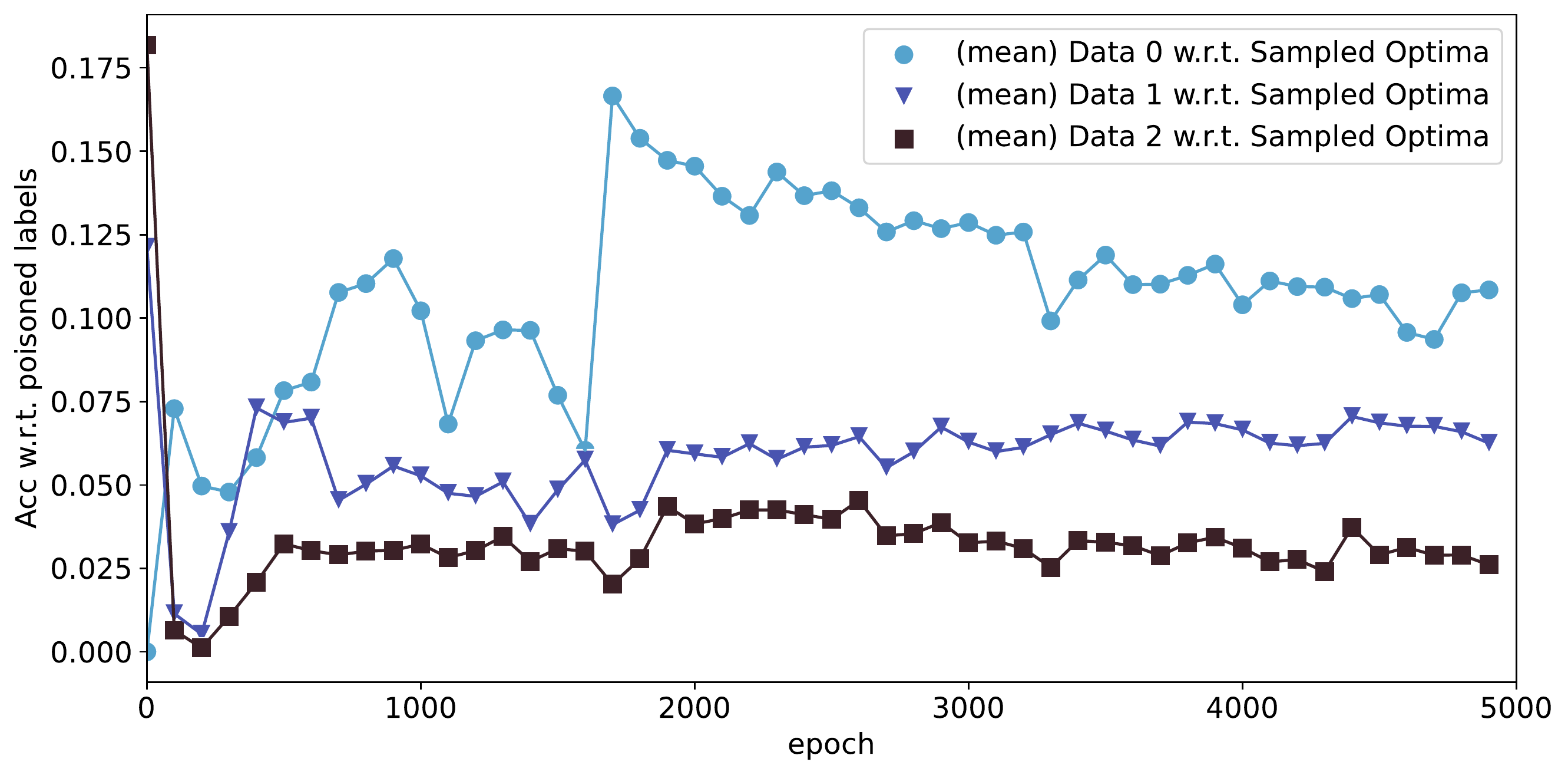}
    \includegraphics[width=0.225\textwidth]{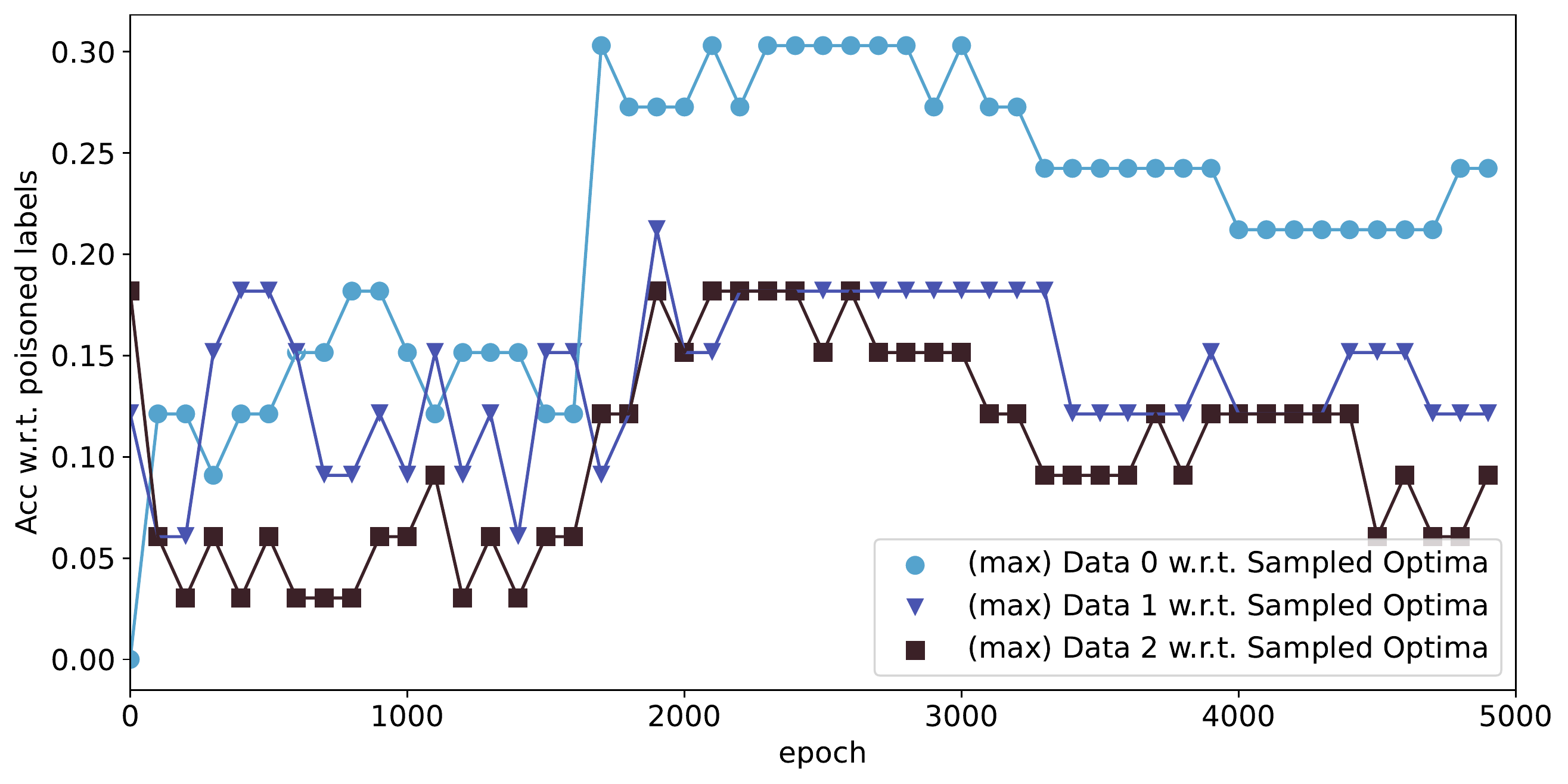}
    \includegraphics[width=0.225\textwidth]{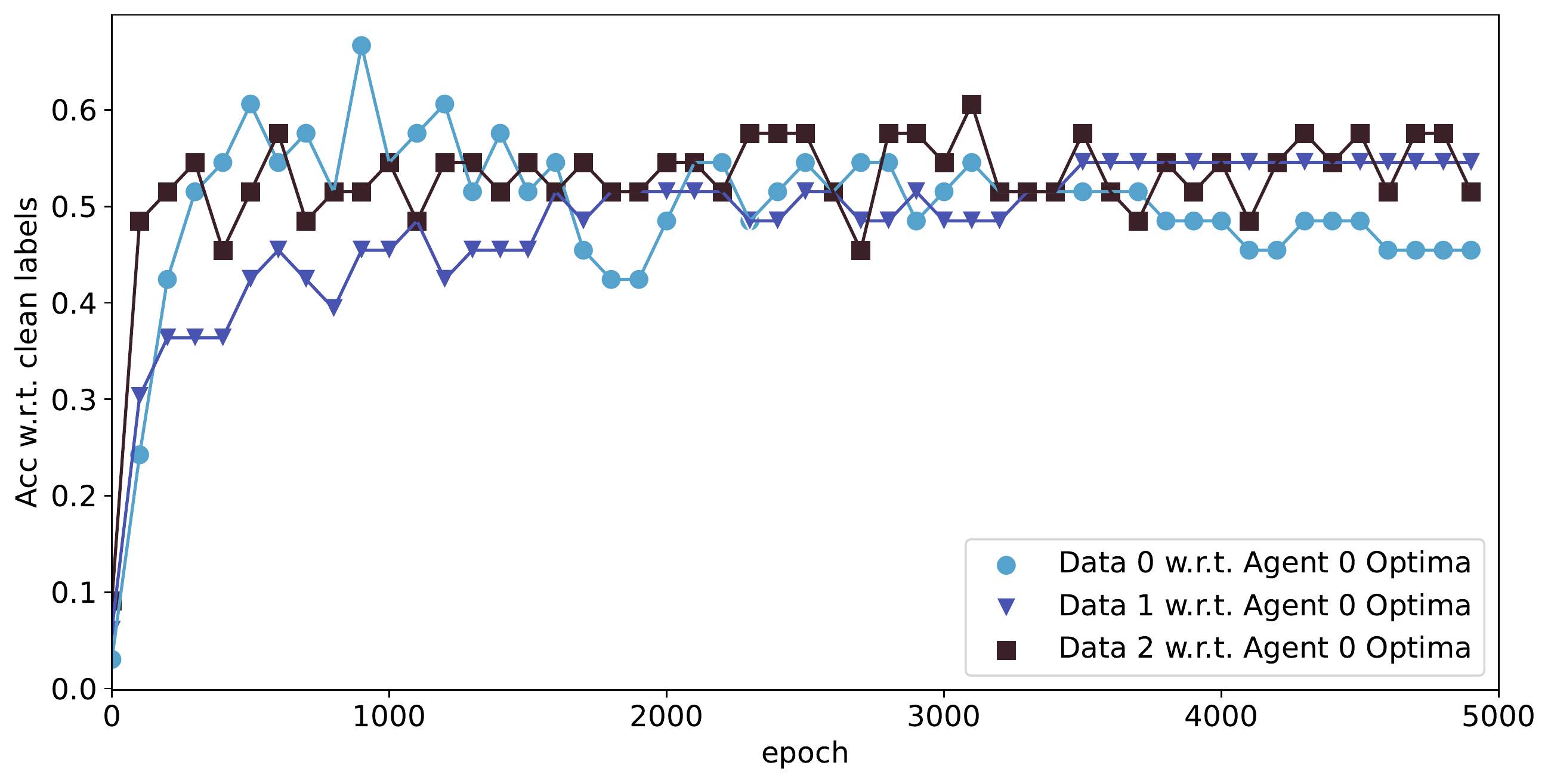}
    \includegraphics[width=0.225\textwidth]{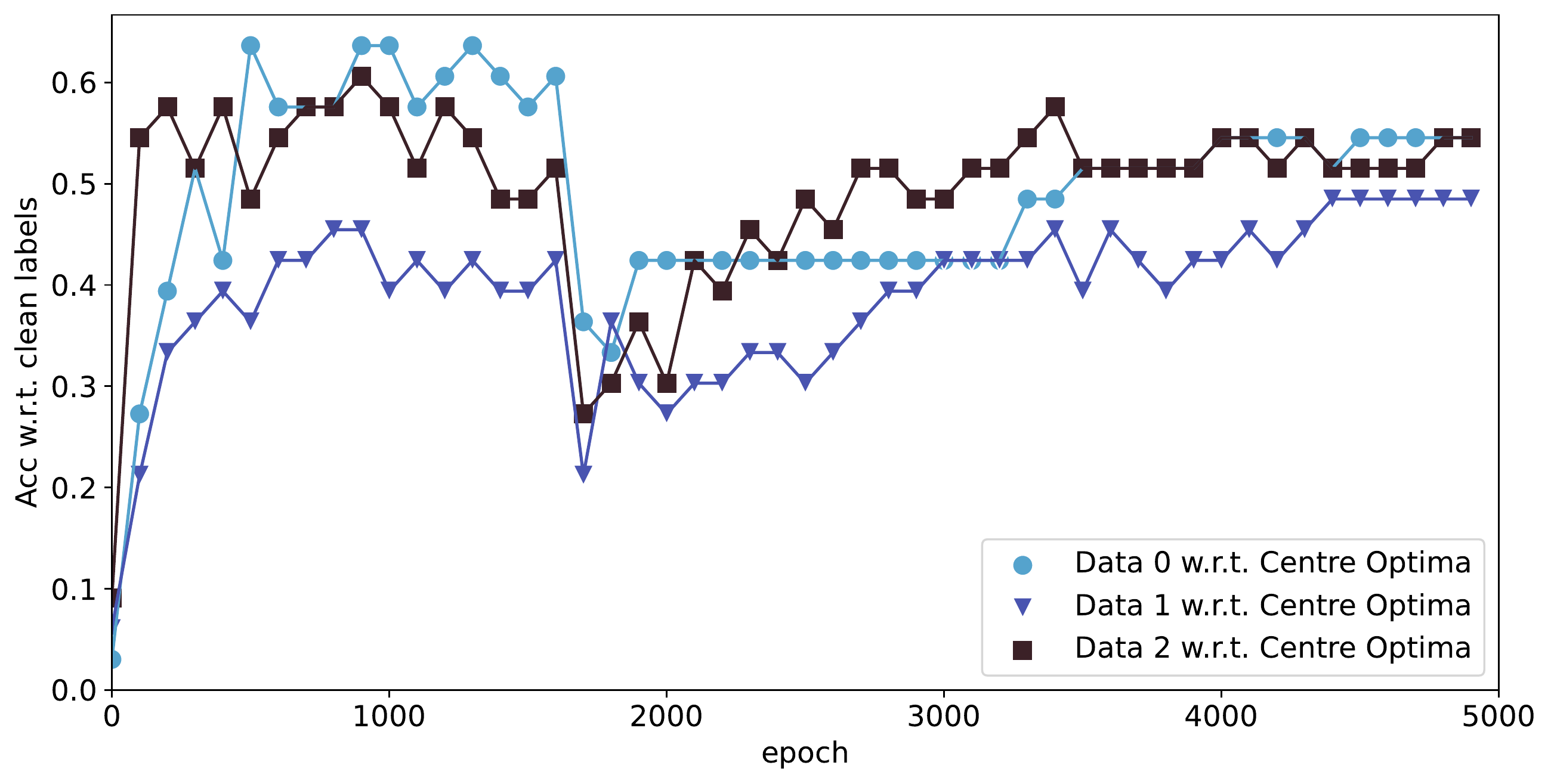}  
    \includegraphics[width=0.225\textwidth]{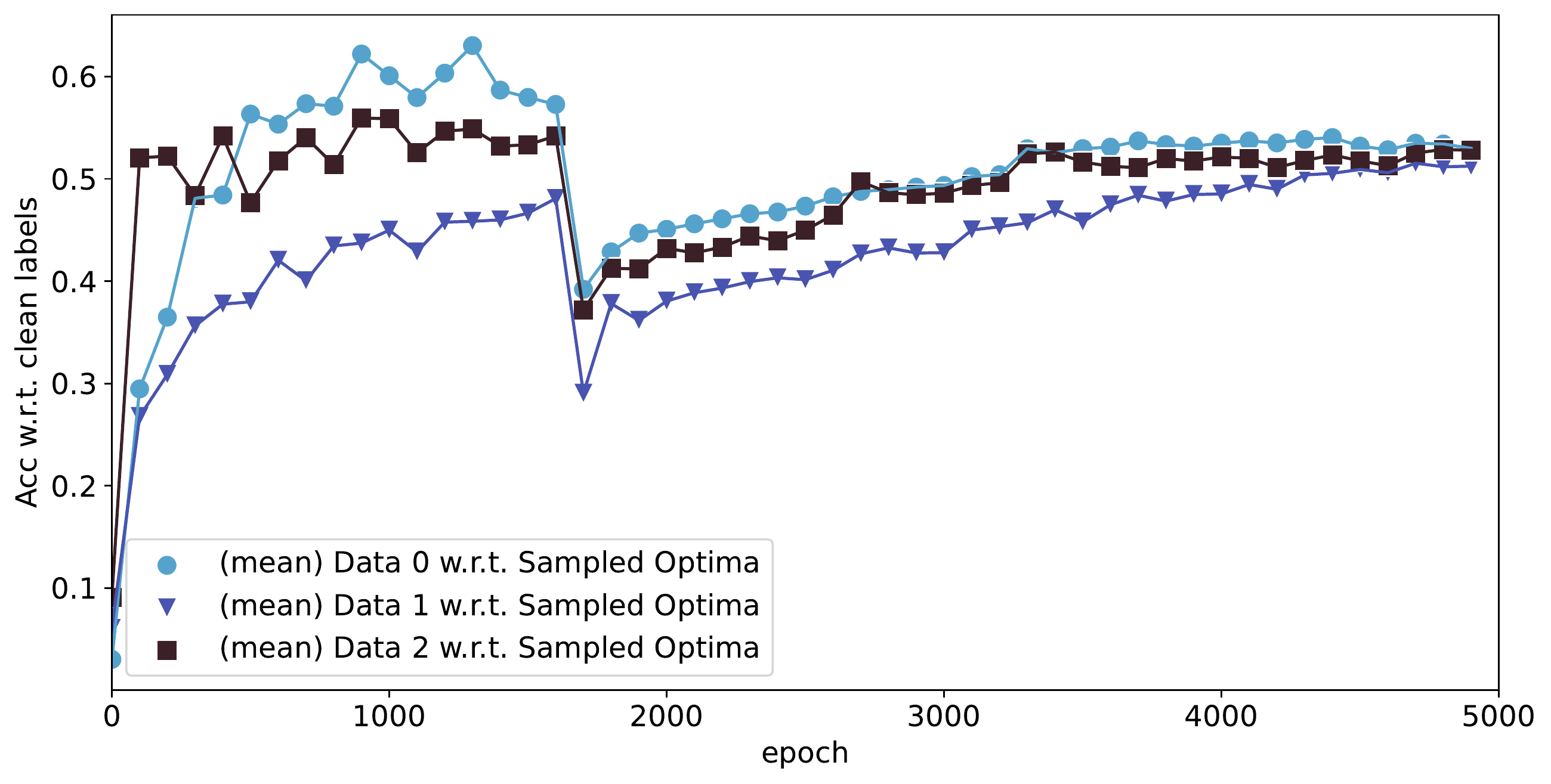}    
    \includegraphics[width=0.225\textwidth]{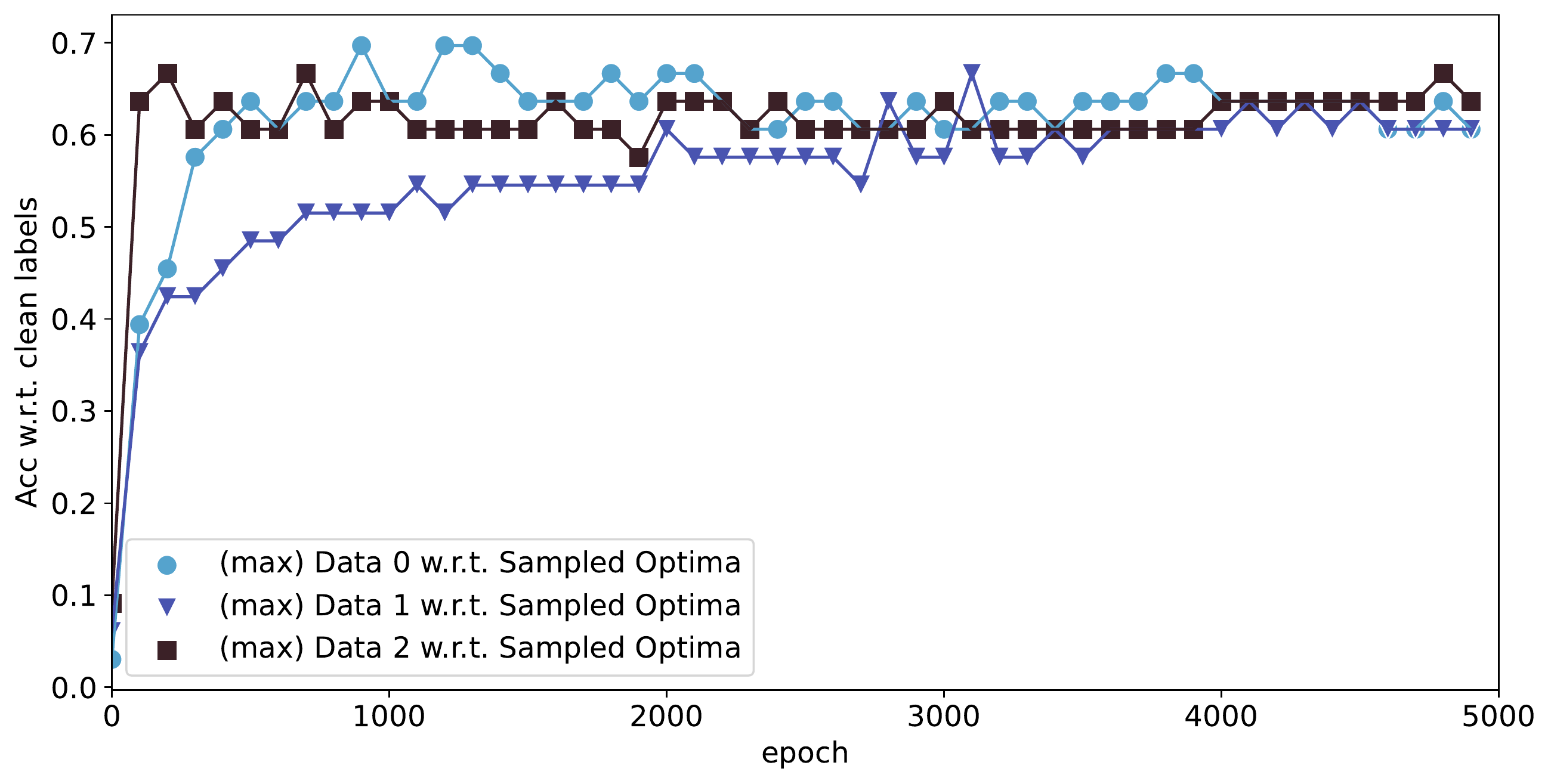}
    \caption{
    Accuracy plots of agents with respect to each other
    }
    \label{fig:max}
\end{figure*}

\subsection{Findings}

\textbf{(Observation 1: Compressed subspace sampling optimally minimizes and maximizes accuracy w.r.t. poisoned and clean labels respectively)}
Across wide variations, we find that sampling with an agent subspace yields the optimal trade-off between maximizing accuracy w.r.t. clean labels and minimizing accuracy w.r.t. poisoned labels.

Summarized in Tables \ref{tab:delta} and \ref{tab:eps}, where we vary the poison parameters $p = \varepsilon$ and number of attackers respectively, the agent subspace method outperforms for most evaluated values.
Evaluating 
on SVNH (digit dataset of 10 classes, 630,420 inputs, 3 colour channels) \citep{37648} (Figure \ref{fig:ablations}b), 
on ResNet-50 \citep{He2015} for the same training configurations (Figure \ref{fig:ablations}c),
and $\alpha=1.0$ stylization with Adaptive Instance Normalization
(AdaIN) \citep{huang2017adain} and the Paintings by Numbers dataset (in-line with Datta \& Shadbolt (2022) \cite{datta2022backdoors} and \citep{geirhos2018imagenettrained}) (Figure \ref{fig:ablations}d), 
we observe that the agent subspace method consistently remains in the top-left region while competing defenses sporadically outperform marginally in different settings.

Compared to single point-estimate defenses (e.g. agent augmentation), subspace sampling methods and other ensembling methods have demonstrated marginal utility from sampling diverse regions of the function space. 
Not only does the agent subspace method outperform all the augmentative and removal strategies, but it also outperforms the agent indexing defense, presumably the upper bound as it is the boundary of the constructed subspace; accuracy w.r.t. clean labels exceed that at this point due to the diversity of functions sampled in ensemble.
Compared to non-agent-dynamics-driven strategies (e.g. ensembles), 
the marginal performance gain is validation that exerting correctness or structure to the subspace for sampling can yield marginal benefits, that subspaces expected to handle multiple distribution shifts should be crafted with multiple distributions in mind, rather than aggregating 1 training set. 

We also observe extended optionality in inference with the agent subspace method. 
Based on Figure \ref{fig:max}, upon convergence we note that evaluating the test-time backdoored set with either the centre of the subspace or ensembling 1000 point-estimates return a similar terminal accuracy w.r.t. clean labels. This implies various storage strategies, ranging from storing a single centre point-estimate, or storing the 3 boundary point-estimates for each attacker. 
We would recommend pursuing further study into crafting efficient heuristics to navigate the subspace: based on the maximum accuracy obtained from a sampled point, we observe that there do exist points within the subspace that offer at least 10\% gains on accuracy w.r.t. clean labels. Possible heuristics could be using meta-data as headers (similar to agent indexing), or using test-time inputs for experience replay.

\begin{table*}[t]
\centering
\parbox{.1625\linewidth}{
\centering

\resizebox{0.1625\textwidth}{!}{

\begin{tabular}{|l|ccccc|}
\multicolumn{1}{c}{}   
\\ \hline
\textit{Num attackers}
\\ \hline\hline

No Defense
\\ \hline

Data Augmentation
\\ \hline

Backdoor Adversarial Training
\\ \hline

Spectral Signatures
\\ \hline

Activation Clustering
\\ \hline

Deep Ensembles
\\ \hline

Fast Geometric Ensembles
\\ \hline\hline

Agent Augmentation
\\ \hline

Agent Indexing
\\ \hline

Agent Subspace
\\ \hline\hline

$\Delta$: random init.
\\ \hline

$\Delta$: train on respective dataset
\\ \hline

$\Delta$:  train sequentially
\\ \hline

$\Delta$:  max. cosine distance
\\ \hline

\end{tabular}
}

}
\parbox{.2125\linewidth}{
\centering

\resizebox{0.2125\textwidth}{!}{

\begin{tabular}{|ccc|}

\hline
\multicolumn{3}{|c|}{Acc w.r.t. poison labels}   
\\ \hline
\textit{2}            
& \textit{5}            
& \textit{10}
\\ \hline\hline

$39.0 \pm 22.0$
&$28.5\pm 21.2$
&$21.0 \pm 19.1$
\\ \hline

$12.8 \pm 0.0$
&$11.0 \pm 1.9$
&$10.1 \pm 7.6$
\\ \hline

$10.0 \pm 0.0$
&$7.7 \pm 3.9$
&$9.9 \pm 3.0$
\\ \hline

$0.0 \pm 0.0$
&$0.1 \pm 0.2$
&$3.7 \pm 7.1$
\\ \hline

$10.7 \pm 3.1$
&$9.44 \pm 4.5$
&$9.1 \pm 7.3$
\\ \hline

$50.0 \pm 50.0$
&$20.0 \pm 40.0$
&$10.0 \pm 30.0$
\\ \hline

$0.0 \pm 0.0$
&$0.0 \pm 0.0$
&$8.0 \pm 7.5$
\\ \hline\hline

$6.0 \pm 3.0$
&$7.5 \pm 3.5$
&$10.0 \pm 6.4$
\\ \hline

$5.5 \pm 2.5$
&$9.0 \pm 3.4$
&$10.4 \pm 5.5$
\\ \hline

$5.132 \pm 0.9$
&$3.41 \pm 2.7$
&$10.1 \pm 3.3$
\\ \hline\hline

$10.1 \pm 6.8$
&$1.34 \pm 1.3$
&$0.67 \pm 1.0$
\\ \hline

$5.79 \pm 2.7$
&$4.83 \pm 3.5$
&$7.91 \pm 6.1$
\\ \hline

$8.71 \pm 0.2$
&$5.7 \pm 1.4$
&$10.4 \pm 7.1$
\\ \hline

$7.72 \pm 1.0$
&$16.1 \pm 14.9$
&$3.94 \pm 2.8$
\\ \hline

\end{tabular}
}

}
\parbox{.2\linewidth}{
\centering

\resizebox{0.2\textwidth}{!}{

\begin{tabular}{|ccc|}

\hline
\multicolumn{3}{|c|}{Acc w.r.t. clean labels}   
\\ \hline
\textit{2}            
& \textit{5}            
& \textit{10}
\\ \hline\hline

$12.5 \pm 1.5$
&$13.5 \pm 4.1$
&$11.5 \pm 7.1$
\\ \hline

$4.65 \pm 2.5$
&$52.6 \pm 3.7$
&$44.0 \pm 7.3$
\\ \hline

$37.5 \pm 4.5$
&$41.5 \pm 3.9$
&$8.5 \pm 1.5$
\\ \hline

$9.25 \pm 1.0$
&$10.6 \pm 1.4$
&$10.9 \pm 2.5$
\\ \hline

$43.0 \pm 0.0$
&$45.0 \pm 1.3$
&$44.8 \pm 3.5$
\\ \hline

$8.0 \pm 0.0$
&$11.0 \pm 1.2$
&$10.0 \pm 6.7$
\\ \hline

$9.0 \pm 0.0$
&$10.0 \pm 2.7$
&$32.0 \pm 11.9$
\\ \hline\hline

$24.5 \pm 3.5$
&$36.5 \pm 7.5$
&$34.9 \pm 4.3$
\\ \hline

$32.0 \pm 2.0$
&$49.5 \pm 5.3$
&$55.8 \pm 5.2$
\\ \hline

{\color{electricultramarine} $48.5 \pm 0.5$}
&{\color{electricultramarine} $55.0 \pm 0.0$}
&{\color{electricultramarine} $62.9 \pm 3.7$}
\\ \hline\hline

$20.8 \pm 0.3$
&$13.5 \pm 1.0$
&$14.9 \pm 2.3$
\\ \hline

$34.1 \pm 2.7$
&$35.8 \pm 3.4$
&$45.8 \pm 11.1$
\\ \hline

$32.5 \pm 0.1$
&$35.4 \pm 4.2$
&$44.0 \pm 10.0$
\\ \hline

$36.3 \pm 0.4$
&$15.1 \pm 1.3$
&$14.7 \pm 3.4$
\\ \hline

\end{tabular}
}

}

\caption{
Defenses: Carefully-configured agent subspace is robust to shifts.
}
\label{tab:delta}
\end{table*}

\begin{table*}[t]
\centering
\parbox{.145\linewidth}{
\centering

\resizebox{0.145\textwidth}{!}{

\begin{tabular}{|l|ccccc|}
\multicolumn{1}{c}{}   
\\ \hline
\textit{Backdoor $\varepsilon$ \& Poison rate}
\\ \hline\hline

Data Augmentation
\\ \hline

Backdoor Adv Training
\\ \hline

Spectral Signatures
\\ \hline

Activation Clustering
\\ \hline

Deep Ensembles
\\ \hline

Fast Geometric Ensembles
\\ \hline\hline

\textbf{Agent Augmentation}
\\ \hline

\textbf{Agent Indexing}
\\ \hline

\textbf{Agent Subspace}
\\ \hline

\end{tabular}
}

}
\parbox{.34\linewidth}{
\centering

\resizebox{0.34\textwidth}{!}{

\begin{tabular}{|ccccc|}

\hline
\multicolumn{5}{|c|}{Acc w.r.t. poison labels}   
\\ \hline
\textit{0.1}            
& \textit{0.2}            
& \textit{0.4}           
& \textit{0.8}
& \textit{1.0}
\\ \hline\hline

$12.1 \pm 6.5$ 
& $12.1 \pm 6.5$ 
& $12.1 \pm 8.5$
& $19.2 \pm 12.5$
& $29.3 \pm 17.5$
\\ \hline

$8.47 \pm 8.8$
& $8.97 \pm 7.6$
& $8.47 \pm 6.1$
& $6.97 \pm 3.0$
& $7.0 \pm 1.4$
\\ \hline

$0.0 \pm 0.0$
& $0.0 \pm 0.0$
& $0.0 \pm 0.0$
& $0.0 \pm 0.0$
& $0.0 \pm 0.0$
\\ \hline

$9.23 \pm 2.0$
& $8.63 \pm 3.3$
& $10.6 \pm 2.8$
& $14.8 \pm 0.9$
& $15.1 \pm 5.7$
\\ \hline

$0.0 \pm 0.0$
& $0.0 \pm 0.0$
& $0.0 \pm 0.0$
& $0.0 \pm 0.0$
& $0.0 \pm 0.0$
\\ \hline

$0.0 \pm 0.0$
& $0.0 \pm 0.0$
& $0.0 \pm 0.0$
& $17.2 \pm 10.3$
& $17.2 \pm 5.7$
\\ \hline\hline

$10.8 \pm 0.5$
& $9.97 \pm 0.6$
& $8.93 \pm 1.7$
& $7.63 \pm 1.7$
& $15.0 \pm 2.9$
\\ \hline

$6.07 \pm 4.9$
& $5.03 \pm 5.1$
& $5.03 \pm 5.1$
& $13.1 \pm 7.6$
& $11.1 \pm 10.0$
\\ \hline

$4.66 \pm 2.8$
& $5.84 \pm 2.2$
& $2.01 \pm 1.7$
& $5.57 \pm 2.7$
& $3.15 \pm 0.2$
\\ \hline

\end{tabular}
}

}
\parbox{.355\linewidth}{
\centering

\resizebox{0.355\textwidth}{!}{

\begin{tabular}{|ccccc|}

\hline
\multicolumn{5}{|c|}{Acc w.r.t. clean labels}   
\\ \hline
\textit{0.1}            
& \textit{0.2}            
& \textit{0.4}           
& \textit{0.8}
& \textit{1.0}
\\ \hline\hline

$31.3 \pm 7.95$
& $30.3 \pm 6.6$
& $31.3 \pm 5.2$
& $33.3 \pm 2.5$
& $28.3 \pm 3.8$
\\ \hline

$19.4 \pm 4.4$
& $21.4 \pm 2.8$
& $20.9 \pm 3.2$
& $20.9 \pm 1.2$
& $21.4 \pm 1.4$
\\ \hline

$9.57 \pm 1.3$
& $9.83 \pm 1.4$
& $9.67 \pm 0.6$
& $9.57 \pm 1.6$
& $9.73 \pm 1.5$
\\ \hline

{\color{electricultramarine} $53.2 \pm 2.1$}
& $48.1 \pm 1.2$
& $42.0 \pm 1.6$
& $34.4 \pm 0.4$
& $32.2 \pm 1.1$
\\ \hline

$11.1 \pm 3.8$
& $11.1 \pm 3.8$
& $11.1 \pm 3.8$
& $11.1 \pm 3.8$
& $11.1 \pm 3.8$
\\ \hline

$10.1 \pm 6.2$
& $10.1 \pm 6.2$
& $10.1 \pm 6.2$
& $21.2 \pm 7.4$
& $17.2 \pm 2.8$
\\ \hline\hline

$44.6 \pm 2.6$
& $40.8 \pm 3.6$
& $34.4 \pm 2.1$
& $37.0 \pm 2.0$
& $33.6 \pm 0.9$
\\ \hline

$43.8 \pm 7.6$
& $46.0 \pm 4.9$
& $46.0 \pm 8.5$
& $44.9 \pm 10.3$
& $37.2 \pm 10.3$
\\ \hline

$50.0 \pm 6.4$
& {\color{electricultramarine} $50.0 \pm 6.4$}
& {\color{electricultramarine} $49.8 \pm 6.1$}
& {\color{electricultramarine} $43.7 \pm 6.1$}
& {\color{electricultramarine} $46.1 \pm 5.1$}
\\ \hline

\end{tabular}
}

}

\caption{
Defenses: Agent subspace is the optimal strategy.
}
\label{tab:eps}
\end{table*}

\textbf{(Observation 2: Distributional robustness play an important role in subspace construction)}
We motivated the 3 defenses based on agent dynamics, which is a manifestation of (joint) distribution shift.
The inclusion of multiple train-time perturbations mapped to different labels is a form of joint label shift, and the stylization of each attacker's private dataset introduces additional domain/texture shift; agent subspacing has exemplified robustness against these different variations of joint distribution shift. 
The marginal utility of the agent subspace methods indicates intrinsic robustness to joint distribution shift.

The performance of this method is implied to be applicable to tasks with \textit{shift-agnosticity} rather than \textit{shift-specificity}.
In the multi-agent backdoor attack, or other task setups where the defender would like to craft a robust adaptation method that performs inference with respect to the source distribution rather than the target distribution, the agent subspace may contain point-estimates that contain specific subnetworks corresponding to specific distributional shift types (e.g. specific tasks or domains), but an expectedly larger proportion of the subspace would contain overlapping, transferable subnetworks. The inference strategies proposed, including inference with the centre or sampling across the entire subspace, empirically performs well when the shifted distributions share one common source distribution, but a heuristic-sampling method could be investigated for searching for specific subnetworks for specific target distributions.

Most work in mode connectivity and subspace construction rely on a single training set, even when showing improvements to adversarial robustness.
Reflecting on single-agent defenses, we train on segregated subpopulations, particularly if the distance between the input distributions in the input space is increasingly distant (e.g. stylization). 
If we train each agent's point-estimate on their private dataset and compress the subspace, the accuracy w.r.t. clean labels is also low, and this may occur because (similarly with the agent-indexing defense):
(i) the \textit{backfiring effect} would occur in that sub-dataset (and not occur if there is only 1 backdoor perturbation type), and
(ii) parameters are expected to learn backdoor subnetworks to minimize loss w.r.t. inputs.

Prior work in mode connectivity and subspace construction make use of random initializations for each point-estimate. This has resulted in lower accuracy in clean labels, and we could attribute this to: (i) additional difficulty in overcoming the marginal parameter space distance between the random initializations, or (ii) the optimization trajectories taken by each training point-estimate has sustained sufficient inertia that it optimizes towards minimizing loss w.r.t. inputs and ignores loss w.r.t. cosine distance. 

\newpage
Wortsman et al. (2021) \cite{wortsman2021learning} constructed a low-loss subspace by jointly training the point-estimates by sampling $\alpha$ coefficients during training to evaluate loss on weighted-average parameters, and additionally regularized for the maximization of cosine distance to include diverse point-estimates. 
Differently, when training on different distributions, we were not able to meaningfully converge the loss of the point-estimates when using $\alpha$ coefficients during training. Maximizing the cosine distance also resulted in reduced accuracy w.r.t. clean labels. We do not use $\alpha$ coefficients during train-time but during test-time for inference, yet the loss on test-time inputs is low. 
We suspect this contradiction is due to the interference between sampled subnetworks: when trained on a single distribution, functionally-different subnetworks mapped to different regions of the function space can stochastically adjust their position in the network given that, though functionally-different, they contribute to minimizing the same input loss (i.e. they are transferable subnetworks). When trained on multiple distributions, the subnetworks interfere with each other, and the subspace cannot be constructed such that a partial subnetwork on one end of the subspace can exist on the other end without 
skewness.

We evaluated the construction of the subspace through sequential point-estimate training, where we optimize the first parameter to its agent-indexed set, then train all subsequent points while minimizing their cosine distances with respect to the first trained point. 
Despite sharing a source distribution, we observe a drop in accuracy w.r.t. clean labels. The implication is that, given an anchoring point-estimate, SGD will not search for an optimal subspace in the parameter space, and instead return interference between minimizing loss w.r.t. their respective dataset or minimizing cosine distance. 
If the former occurs, it is implied the subspace is neither compressed nor unlikely to carry transferable subnetworks of the source distribution.
If the latter occurs, then the subsequently trained points will implicitly learn subnetworks skewed towards the first point, even though it has not encountered those backdoor perturbations in its own training set.
During training in parallel, we also need to mitigate this skewness of subnetwork insertion: if the cosine distance during the initial epochs are too high and diverge, then the loss of the individual parameters may not converge, hence we set the regularization coefficient lower to reduce the loss terms imbalance.

A feature of this work is the intrinsic re-calibration taking place per epoch. 
Opposing the expectation of a continuously-decreasing trend in the cosine distance, we see a double/multiple descent phenomenon occurring, where while optimizing loss w.r.t. inputs the point-estimates are finding optimal subspace regions in the parameter space such that the cosine distance can continue to decrease, and the search for this region is what yields a temporary increase in cosine distance. 
We naturally yield a compressed low-loss subspace without the use of substantial construction tweaks such as alpha coefficients. 
If we stopped training at epoch 1000, we would retain skew and this would benefit some models, but if we continue training we find a terminal equilibrium accuracy w.r.t. clean labels that matches the mean at epoch 1000 but lower variance.


\section{Conclusion}

This work contributed a robust, practical defense against multi-agent backdoor attacks, specifically maximizing the accuracy w.r.t. clean labels which was harmed during backfire. We also contributed novel methodology using compressed low-loss subspaces to mitigate distribution shifts.


{\small
\bibliographystyle{ieee_fullname}
\bibliography{main}

\begin{thebibliography}{10}\itemsep=-1pt

\bibitem{pmlr-v108-bagdasaryan20a}
Eugene Bagdasaryan, Andreas Veit, Yiqing Hua, Deborah Estrin, and Vitaly
  Shmatikov.
\newblock How to backdoor federated learning.
\newblock In Silvia Chiappa and Roberto Calandra, editors, {\em Proceedings of
  the Twenty Third International Conference on Artificial Intelligence and
  Statistics}, volume 108 of {\em Proceedings of Machine Learning Research},
  pages 2938--2948. PMLR, 26--28 Aug 2020.

\bibitem{9414862}
Eitan Borgnia, Valeriia Cherepanova, Liam Fowl, Amin Ghiasi, Jonas Geiping,
  Micah Goldblum, Tom Goldstein, and Arjun Gupta.
\newblock Strong data augmentation sanitizes poisoning and backdoor attacks
  without an accuracy tradeoff.
\newblock In {\em ICASSP 2021 - 2021 IEEE International Conference on
  Acoustics, Speech and Signal Processing (ICASSP)}, pages 3855--3859, 2021.

\bibitem{chen2018detecting}
Bryant Chen, Wilka Carvalho, Nathalie Baracaldo, Heiko Ludwig, Benjamin
  Edwards, Taesung Lee, Ian Molloy, and Biplav Srivastava.
\newblock Detecting backdoor attacks on deep neural networks by activation
  clustering, 2018.

\bibitem{chen2017targeted}
Xinyun Chen, Chang Liu, Bo Li, Kimberly Lu, and Dawn Song.
\newblock Targeted backdoor attacks on deep learning systems using data
  poisoning, 2017.

\bibitem{CHEN2021100002}
Zheyi Chen, Pu Tian, Weixian Liao, and Wei Yu.
\newblock Towards multi-party targeted model poisoning attacks against
  federated learning systems.
\newblock {\em High-Confidence Computing}, page 100002, 2021.

\bibitem{datta2022backdoors}
Siddhartha Datta and Nigel Shadbolt.
\newblock Backdoors stuck at the frontdoor: Multi-agent backdoor attacks that
  backfire.
\newblock In {\em International Conference on Learning Representations
  Workshop: Gamification and Multiagent Solutions}, 2022.

\bibitem{dattaSelf}
Siddhartha Datta and Nigel Shadbolt.
\newblock Hiding behind backdoors: Self-obfuscation against generative models,
  2022.

\bibitem{https://doi.org/10.48550/arxiv.2205.09891}
Siddhartha Datta and Nigel Shadbolt.
\newblock Interpolating compressed parameter subspaces, 2022.

\bibitem{draxler2019essentially}
Felix Draxler, Kambis Veschgini, Manfred Salmhofer, and Fred~A. Hamprecht.
\newblock Essentially no barriers in neural network energy landscape, 2019.

\bibitem{247652}
Minghong Fang, Xiaoyu Cao, Jinyuan Jia, and Neil Gong.
\newblock Local model poisoning attacks to byzantine-robust federated learning.
\newblock In {\em 29th {USENIX} Security Symposium ({USENIX} Security 20)},
  pages 1605--1622. {USENIX} Association, Aug. 2020.

\bibitem{fort2020deep}
Stanislav Fort, Huiyi Hu, and Balaji Lakshminarayanan.
\newblock Deep ensembles: A loss landscape perspective, 2020.

\bibitem{fort2019large}
Stanislav Fort and Stanislaw Jastrzebski.
\newblock Large scale structure of neural network loss landscapes, 2019.

\bibitem{frankle2020linear}
Jonathan Frankle, Gintare~Karolina Dziugaite, Daniel~M. Roy, and Michael
  Carbin.
\newblock Linear mode connectivity and the lottery ticket hypothesis, 2020.

\bibitem{gao2020backdoor}
Yansong Gao, Bao~Gia Doan, Zhi Zhang, Siqi Ma, Jiliang Zhang, Anmin Fu, Surya
  Nepal, and Hyoungshick Kim.
\newblock Backdoor attacks and countermeasures on deep learning: A
  comprehensive review, 2020.

\bibitem{garipov2018loss}
Timur Garipov, Pavel Izmailov, Dmitrii Podoprikhin, Dmitry Vetrov, and
  Andrew~Gordon Wilson.
\newblock Loss surfaces, mode connectivity, and fast ensembling of dnns, 2018.

\bibitem{geiping2021doesnt}
Jonas Geiping, Liam Fowl, Gowthami Somepalli, Micah Goldblum, Michael Moeller,
  and Tom Goldstein.
\newblock What doesn't kill you makes you robust(er): Adversarial training
  against poisons and backdoors, 2021.

\bibitem{geirhos2018imagenettrained}
Robert Geirhos, Patricia Rubisch, Claudio Michaelis, Matthias Bethge, Felix~A.
  Wichmann, and Wieland Brendel.
\newblock Imagenet-trained {CNN}s are biased towards texture; increasing shape
  bias improves accuracy and robustness.
\newblock In {\em International Conference on Learning Representations}, 2019.

\bibitem{goodfellow2015qualitatively}
Ian~J. Goodfellow, Oriol Vinyals, and Andrew~M. Saxe.
\newblock Qualitatively characterizing neural network optimization problems,
  2015.

\bibitem{gu2019badnets}
Tianyu Gu, Brendan Dolan-Gavitt, and Siddharth Garg.
\newblock Badnets: Identifying vulnerabilities in the machine learning model
  supply chain, 2019.

\bibitem{8685687}
Tianyu Gu, Kang Liu, Brendan Dolan-Gavitt, and Siddharth Garg.
\newblock Badnets: Evaluating backdooring attacks on deep neural networks.
\newblock {\em IEEE Access}, 7:47230--47244, 2019.

\bibitem{NEURIPS2018_331316d4}
Jamie Hayes and Olga Ohrimenko.
\newblock Contamination attacks and mitigation in multi-party machine learning.
\newblock In S. Bengio, H. Wallach, H. Larochelle, K. Grauman, N. Cesa-Bianchi,
  and R. Garnett, editors, {\em Advances in Neural Information Processing
  Systems}, volume~31. Curran Associates, Inc., 2018.

\bibitem{He2015}
Kaiming He, Xiangyu Zhang, Shaoqing Ren, and Jian Sun.
\newblock Deep residual learning for image recognition.
\newblock {\em arXiv preprint arXiv:1512.03385}, 2015.

\bibitem{huang2020dynamic}
Anbu Huang.
\newblock Dynamic backdoor attacks against federated learning, 2020.

\bibitem{huang2017adain}
Xun Huang and Serge Belongie.
\newblock Arbitrary style transfer in real-time with adaptive instance
  normalization.
\newblock In {\em ICCV}, 2017.

\bibitem{izmailov2019subspace}
Pavel Izmailov, Wesley~J. Maddox, Polina Kirichenko, Timur Garipov, Dmitry
  Vetrov, and Andrew~Gordon Wilson.
\newblock Subspace inference for bayesian deep learning, 2019.

\bibitem{izmailov2019averaging}
Pavel Izmailov, Dmitrii Podoprikhin, Timur Garipov, Dmitry Vetrov, and
  Andrew~Gordon Wilson.
\newblock Averaging weights leads to wider optima and better generalization,
  2019.

\bibitem{krizhevsky2009learning}
Alex {Krizhevsky}.
\newblock Learning multiple layers of features from tiny images.
\newblock 2009.

\bibitem{lakshminarayanan2017simple}
Balaji Lakshminarayanan, Alexander Pritzel, and Charles Blundell.
\newblock Simple and scalable predictive uncertainty estimation using deep
  ensembles, 2017.

\bibitem{li2018measuring}
Chunyuan Li, Heerad Farkhoor, Rosanne Liu, and Jason Yosinski.
\newblock Measuring the intrinsic dimension of objective landscapes, 2018.

\bibitem{li2021backdoor}
Yiming Li, Baoyuan Wu, Yong Jiang, Zhifeng Li, and Shu-Tao Xia.
\newblock Backdoor learning: A survey, 2021.

\bibitem{9230411}
G. {Lovisotto}, S. {Eberz}, and I. {Martinovic}.
\newblock Biometric backdoors: A poisoning attack against unsupervised template
  updating.
\newblock In {\em 2020 IEEE European Symposium on Security and Privacy
  (EuroS\&P)}, pages 184--197, 2020.

\bibitem{mahloujifar2018multiparty}
Saeed Mahloujifar, Mohammad Mahmoody, and Ameer Mohammed.
\newblock Multi-party poisoning through generalized $p$-tampering, 2018.

\bibitem{pmlr-v97-mahloujifar19a}
Saeed Mahloujifar, Mohammad Mahmoody, and Ameer Mohammed.
\newblock Universal multi-party poisoning attacks.
\newblock In Kamalika Chaudhuri and Ruslan Salakhutdinov, editors, {\em
  Proceedings of the 36th International Conference on Machine Learning},
  volume~97 of {\em Proceedings of Machine Learning Research}, pages
  4274--4283. PMLR, 09--15 Jun 2019.

\bibitem{37648}
Yuval Netzer, Tao Wang, Adam Coates, Alessandro Bissacco, Bo Wu, and Andrew~Y.
  Ng.
\newblock Reading digits in natural images with unsupervised feature learning.
\newblock In {\em NIPS Workshop on Deep Learning and Unsupervised Feature
  Learning 2011}, 2011.

\bibitem{nunez2021lcs}
Elvis Nunez, Maxwell Horton, Anish Prabhu, Anurag Ranjan, Ali Farhadi, and
  Mohammad Rastegari.
\newblock Lcs: Learning compressible subspaces for adaptive network compression
  at inference time, 2021.

\bibitem{ratzlaff2020hypergan}
Neale Ratzlaff and Li Fuxin.
\newblock Hypergan: A generative model for diverse, performant neural networks,
  2020.

\bibitem{saha2020hidden}
Aniruddha Saha, Akshayvarun Subramanya, and Hamed Pirsiavash.
\newblock Hidden trigger backdoor attacks.
\newblock In {\em Proceedings of the AAAI Conference on Artificial
  Intelligence}, volume~34, pages 11957--11965, 2020.

\bibitem{shafahi2018poison}
Ali Shafahi, W~Ronny Huang, Mahyar Najibi, Octavian Suciu, Christoph Studer,
  Tudor Dumitras, and Tom Goldstein.
\newblock Poison frogs! targeted clean-label poisoning attacks on neural
  networks.
\newblock {\em arXiv preprint arXiv:1804.00792}, 2018.

\bibitem{48698}
Ananda~Theertha Suresh, Brendan McMahan, Peter Kairouz, and Ziteng Sun.
\newblock Can you really backdoor federated learning?
\newblock 2019.

\bibitem{NEURIPS2018_280cf18b}
Brandon Tran, Jerry Li, and Aleksander Madry.
\newblock Spectral signatures in backdoor attacks.
\newblock In S. Bengio, H. Wallach, H. Larochelle, K. Grauman, N. Cesa-Bianchi,
  and R. Garnett, editors, {\em Advances in Neural Information Processing
  Systems}, volume~31. Curran Associates, Inc., 2018.

\bibitem{vonoswald2021neural}
Johannes von Oswald, Seijin Kobayashi, João Sacramento, Alexander Meulemans,
  Christian Henning, and Benjamin~F. Grewe.
\newblock Neural networks with late-phase weights, 2021.

\bibitem{NEURIPS2020_b8ffa41d}
Hongyi Wang, Kartik Sreenivasan, Shashank Rajput, Harit Vishwakarma, Saurabh
  Agarwal, Jy-yong Sohn, Kangwook Lee, and Dimitris Papailiopoulos.
\newblock Attack of the tails: Yes, you really can backdoor federated learning.
\newblock In H. Larochelle, M. Ranzato, R. Hadsell, M.~F. Balcan, and H. Lin,
  editors, {\em Advances in Neural Information Processing Systems}, volume~33,
  pages 16070--16084. Curran Associates, Inc., 2020.

\bibitem{wortsman2021learning}
Mitchell Wortsman, Maxwell Horton, Carlos Guestrin, Ali Farhadi, and Mohammad
  Rastegari.
\newblock Learning neural network subspaces, 2021.

\bibitem{wortsman2020supermasks}
Mitchell Wortsman, Vivek Ramanujan, Rosanne Liu, Aniruddha Kembhavi, Mohammad
  Rastegari, Jason Yosinski, and Ali Farhadi.
\newblock Supermasks in superposition, 2020.

\bibitem{Xie2020DBA}
Chulin Xie, Keli Huang, Pin-Yu Chen, and Bo Li.
\newblock Dba: Distributed backdoor attacks against federated learning.
\newblock In {\em International Conference on Learning Representations}, 2020.

\bibitem{Yun_2019_ICCV}
Sangdoo Yun, Dongyoon Han, Seong~Joon Oh, Sanghyuk Chun, Junsuk Choe, and
  Youngjoon Yoo.
\newblock Cutmix: Regularization strategy to train strong classifiers with
  localizable features.
\newblock In {\em Proceedings of the IEEE/CVF International Conference on
  Computer Vision (ICCV)}, October 2019.

\bibitem{zhang2018mixup}
Hongyi Zhang, Moustapha Cisse, Yann~N. Dauphin, and David Lopez-Paz.
\newblock mixup: Beyond empirical risk minimization, 2018.

\bibitem{zhu2019transferable}
Chen Zhu, W~Ronny Huang, Hengduo Li, Gavin Taylor, Christoph Studer, and Tom
  Goldstein.
\newblock Transferable clean-label poisoning attacks on deep neural nets.
\newblock In {\em International Conference on Machine Learning}, pages
  7614--7623. PMLR, 2019.

\end{thebibliography}
}

\end{document}